\documentclass[11pt]{article}

\usepackage[preprint]{acl}

\usepackage{times}
\usepackage{latexsym}
\usepackage{booktabs}
\usepackage{xcolor}
\usepackage{pifont}
\usepackage{amsmath}
\usepackage[capitalize,noabbrev,nameinlink]{cleveref}
\usepackage{indentfirst}
\usepackage{multirow}
\usepackage{graphicx}
\usepackage{subcaption}
\usepackage{float}
\usepackage{xcolor}
\usepackage[T1]{fontenc}

\usepackage[utf8]{inputenc}

\usepackage{microtype}

\usepackage{inconsolata}

\usepackage{graphicx}

\newcommand{\greencheck}{\textcolor{green!70!black}{\ding{51}}}

\title{Mid-Think: Training-Free Intermediate-Budget Reasoning via\\ Token-Level Triggers}

\author{\textbf{Wang Yang$^{1*}$, Shouren Wang$^{1*}$, Debargha Ganguly$^1$, Xinpeng Li$^1$}\\ \textbf{Chaoda Song$^1$, Vikash Singh$^1$, Vipin Chaudhary$^1$, Xiaotian Han$^1$}\\
$^1$Case Western Reserve University \\$^*$Equal contribution\\
\texttt{\{wxy320,dxg512,sxw992,vxs465,vipin,xhan\}@case.edu}
}

\begin{document}
\maketitle
\begin{abstract}


Hybrid reasoning language models are commonly controlled through high-level Think/No-think instructions to regulate reasoning behavior, yet we found that such mode switching is largely driven by a small set of trigger tokens rather than the instructions themselves. Through attention analysis and controlled prompting experiments, we show that a leading ``Okay'' token induces reasoning behavior, while the newline pattern following ``</think>'' suppresses it. Based on this observation, we propose \emph{Mid-Think}, a simple training-free prompting format that combines these triggers to achieve intermediate-budget reasoning, consistently outperforming fixed-token and prompt-based baselines in terms of the accuracy-length trade-off. Furthermore, applying Mid-Think to RL training after SFT reduces training time by approximately 15\% while improving final performance of Qwen3-8B on AIME from 69.8\% to 72.4\% and on GPQA from 58.5\% to 61.1\%, demonstrating its effectiveness for both inference-time control and RL-based reasoning training. Our code is available at \url{https://github.com/uservan/Mid-Think}.  

\end{abstract}

\section{Introduction}


Large language models exhibit a variety of emergent phenomena~\cite{sun2024massive,robinson2024sparse}, many of which have been actively exploited to improve model. The work of Attention Sink~\cite{xiao2023efficient, gu2024attention} has been leveraged to accelerate long-context inference. Explicit </think> tags are used to expose intermediate reasoning and enable hybrid thinking behaviors~\cite{fang2025thinkless,yang2025qwen3}. Token-level cues such as \emph{``wait''} and \emph{``alternatively''} have been analyzed to regulate reasoning probability and entropy, forming the methods like No-Wait~\cite{wang2025wait} and SpecExit~\cite{yang2025specexit}. Speculative Thinking~\cite{yang2025speculative} exploits structural patterns such as \texttt{\textbackslash n\textbackslash n} to coordinate cooperation between large and small reasoning models.

\begin{figure}[t]
    \centering
    \begin{subfigure}[t]{0.9\linewidth}
        \centering
        \includegraphics[width=\linewidth]{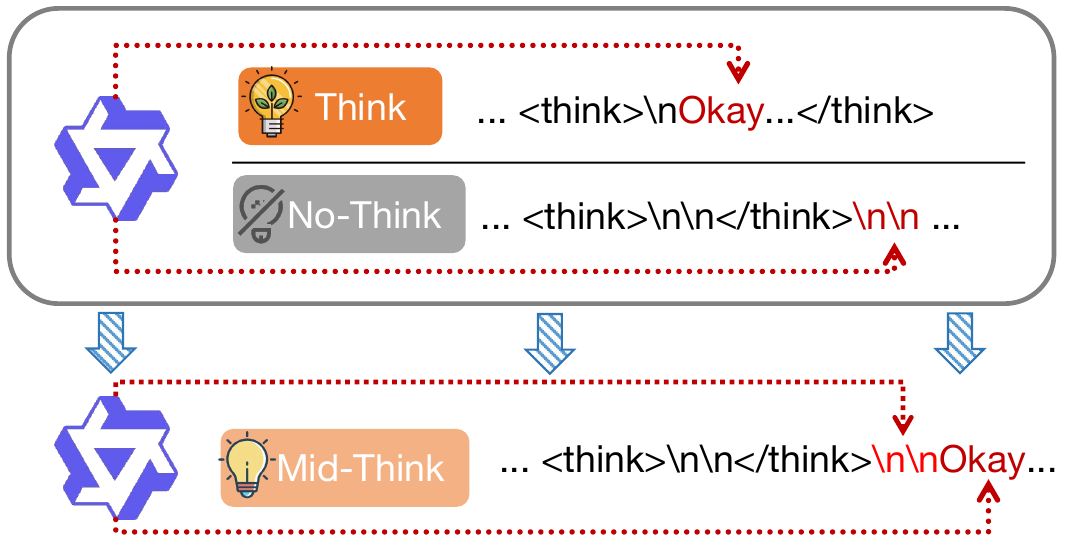}
        \vspace{-18pt}
        \caption{Overview}
    \end{subfigure}
    \begin{subfigure}[t]{0.47\linewidth}
        \centering
        \includegraphics[width=\linewidth]{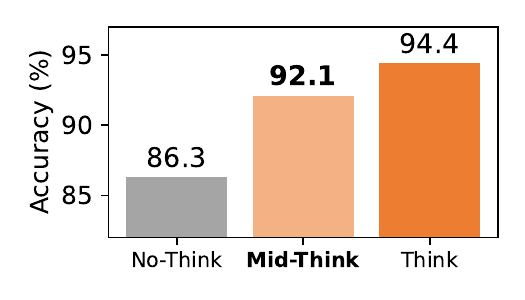}
        \vspace{-18pt}
        \caption{Accuracy}
    \end{subfigure}
    \hfill
    \begin{subfigure}[t]{0.47\linewidth}
        \centering
        \includegraphics[width=\linewidth]{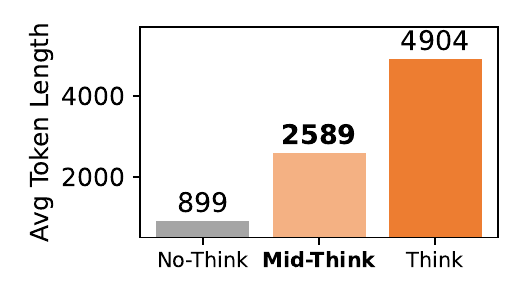}
        \vspace{-18pt}
        \caption{Average Length}
    \end{subfigure}
    \vspace{-6pt}
    \caption{
        Illustration of Mid-Think and its performance on MATH500. (a) Overview of Mid-Think comparing with Think and No-Think.
        In the Think mode, subsequent tokens primarily attend to the cue token ``Okay'',
        while in the No-think mode, generated tokens focus on the newline following the </think> marker (i.e., the \texttt{\textbackslash n\textbackslash n} token).
        Mid-Think combines both cues into a unified prompting format to induce intermediate reasoning behavior. (b) Accuracy and (c) average output length under No-think, Think, and Mid-Think settings.
        Mid-Think achieves intermediate-budget reasoning without additional training, retaining most accuracy gains while significantly reducing generation length.
        }
    \label{fig:overview}
\end{figure}


As shown in \Cref{fig:overview}, we find that \textbf{the Think and No-think behaviors of reasoning models are largely governed by a small number of trigger tokens}. Using Qwen3-8B as a representative example, we compute the average attention from generated tokens to each prompt token. The analysis reveals that only a few tokens consistently attracts substantially higher attention than other prompt components. Specifically, in the Think mode, reasoning behavior is dominated by the ``Okay'' token immediately following the <think> tag, whereas in the No-think mode, the model primarily attends to the newline pattern after the </think> tag.

To further verify this phenomenon, we design multiple prompting formats and evaluate them on MATH500, AIME, and GPQA~\cite{rein2024gpqa}. We observe that prompts containing a leading ``Okay'' token consistently induce reasoning behavior, achieving accuracy and wait count comparable to the standard Think mode. In contrast, </think>+\textbackslash n\textbackslash n pattern significantly reduces accuracy and wait count, yielding behavior closer to No-think regime. Together, it demonstrate that reasoning and non-reasoning behaviors are not determined by high-level instructions, but are instead dominated by a small set of token-level triggers.

Motivated by this observation, \textbf{we propose a new format, termed Mid-Think (<think>\textbackslash n\textbackslash n</think>\textbackslash n\textbackslash n<reason>\textbackslash nOkay\ldots)}, which enables intermediate-budget reasoning without any additional training. By explicitly integrating both reasoning-activating and reasoning-suppressing cues, Mid-Think induces a balanced reasoning behavior between the standard Think and No-Think modes. The comparison results of these modes are shown in \Cref{fig:overview}.

Empirically, we find that Mid-Think consistently achieves performance comparable to, and in some cases better than, fixed intermediate budgets, effectively lying on or beyond the Pareto frontier between accuracy and output length, Mid-Think outperforms existing training-free approaches proposed in Qwen3, including fixed-token budgets and prompt-based budget control, demonstrating a more reliable and fine-grained mechanism for reasoning budget regulation.

Finally, we apply Mid-Think to RL training after SFT and observe consistent improvements in both efficiency and performance. Compared to standard Think-mode training, Mid-Think significantly reduces RL training time while achieving better post-training results. For Qwen3-8B, a model trained with Mid-Think attains higher accuracy when evaluated in the Think mode on AIME (72.4\% vs.\ 69.8\%), while reducing training time from 54 hours to 46 hours, corresponding to an efficiency gain of approximately 15\%.

\section{Motivation: Reasoning Is Governed by a Few Tokens}


This section presents an empirical observation of an \emph{"overfitting"} phenomenon in the reasoning processes. We first analyze the homogenization of patterns in reasoning modes and datasets. Then from an attention-based perspective, we observe that models tend to focus on a small set of specific tokens: the reasoning mode is largely anchored to the lexical cue ``Okay'', while the non-reasoning mode is primarily associated with \texttt{``\textbackslash n\textbackslash n''} following the </think> tag. Finally, we design controlled experiments with different reasoning formats to systematically validate this phenomenon.

\begin{table}[t]
\centering
\small
\setlength{\tabcolsep}{8pt}
\begin{tabular}{lccc}
\toprule
\textbf{Token} 
& \textbf{DeepSeek} 
& \textbf{Qwen3} 
& \textbf{OpenR1-Math} \\
\midrule
<think>   
& \greencheck & \greencheck & \greencheck \\
</think>  
& \greencheck & \greencheck & \greencheck \\
\texttt{Okay}      
& \greencheck & \greencheck & \greencheck \\
\bottomrule
\end{tabular}
\vspace{-8pt}
\caption{Occurrence of explicit reasoning-related tokens in the outputs of reasoning models and training datasets. OpenR1-Math means the reasoning training dataset of open-r1/OpenR1-Math-220k}
\label{tab:token-model-three}
\end{table}

\begin{figure*}[ht]
    \centering
    \includegraphics[width=1.0\linewidth]{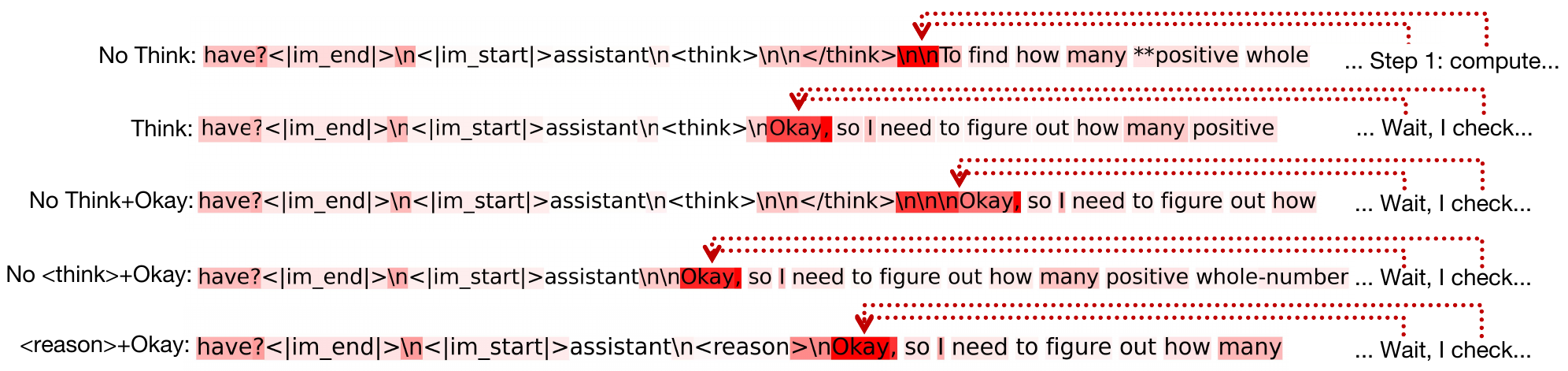}
    \vspace{-25pt}
    \caption{
    Average attention from generated tokens to different opening tokens under five generation modes. Darker red indicates higher attention received from subsequent tokens. When “Okay” appears at the beginning, the model produces an explicit reasoning process ("wait", "alternatively", etc), and attention is primarily concentrated on “Okay”. In the No-Think mode, the \texttt{\textbackslash n\textbackslash n} following </think> absorbs most of the attention mass.
    }
    \label{fig:attention_modes}
\end{figure*}

\begin{table*}[ht]
\centering
\small
\setlength{\tabcolsep}{5pt}
\begin{tabular}{c c cccccc}
\toprule
\multirow{2}{*}{\textbf{Mode}} &
\multirow{2}{*}{\textbf{Format}} &
\multicolumn{2}{c}{\textbf{MATH500}} &
\multicolumn{2}{c}{\textbf{AIME}} &
\multicolumn{2}{c}{\textbf{GPQA}} \\
\cmidrule(lr){3-4} \cmidrule(lr){5-6} \cmidrule(lr){7-8}
& & Acc (\%) & Wait & Acc (\%) & Wait & Acc (\%) & Wait \\
\midrule
\textbf{No-think}
& \texttt{<think>\textbackslash n\textbackslash n</think>\textbackslash n\textbackslash n}
& 83.2 &  2082
& 21.3 &   447
& 37.5 &   459 \\

\textbf{Think}
& \texttt{<think>\textbackslash nOkay}
& 94.6 & 81367
& 71.6 & 40342
& 60.9 & 71817 \\

\textbf{No Think + Okay}
& \texttt{<think>\textbackslash n\textbackslash n</think>\textbackslash n\textbackslash nOkay}
& 92.3 & 33724
& 55.3 & 19360
& 47.3 &  56177 \\

\textbf{No <think> + Okay}
& \texttt{\textbackslash n\textbackslash nOkay}
& 94.1 & 80272
& 72.7 & 40407
& 59.3 & 71914 \\

\textbf{<reason> + Okay}
& \texttt{<reason>\textbackslash nOkay}
& 94.0 & 81075
& 72.7 & 38067
& 56.6 & 66105 \\
\bottomrule
\end{tabular}
\vspace{-8pt}
\caption{
Formats and performance statistics under five different generation modes of Qwen3-8B. The Format column specifies the exact prompting patterns used to trigger different behaviors. We report accuracy on MATH500, AIME, and GPQA, together with the corresponding wait counts. When the prompt begins with “Okay”, both accuracy and wait statistics closely match those of the Think mode.
}
\label{tab:format_mode_dataset}
\end{table*}

\begin{table*}[ht]
\centering
\small
\setlength{\tabcolsep}{10pt}
\begin{tabular}{c p{0.78\textwidth}}
\toprule
\textbf{Mode} & \textbf{Format (literal \texttt{\textbackslash n} denotes newlines)} \\
\midrule
\textbf{Think}
& \texttt{<|im\_start|>user\{query\}<|im\_end|><|im\_start|>assistant\textcolor{red}{<think>}\{thinking\_content\} \textcolor{red}{</think>}\{response\}<|im\_end|>} \\

\textbf{No-Think}
& \texttt{<|im\_start|>user\{query\}<|im\_end|><|im\_start|>assistant\textcolor{red}{<think>\textbackslash n\textbackslash n</think>} \{response\}<|im\_end|>}\\

\textbf{Mid-Think}
& \texttt{<|im\_start|>user\{query\}<|im\_end|><|im\_start|>assistant<think>\textbackslash n\textbackslash n</think> \textbackslash n\textbackslash n\textcolor{red}{<reason>\textbackslash nOkay}...\{thinking\_content\}\textcolor{red}{</reason>}\{response\}<|im\_end|>} \\
\bottomrule
\end{tabular}
\vspace{-8pt}
\caption{Prompt formats under three generation modes. Mid-Think combines key tokens from Think and No-Think (e.g., “Okay” and \texttt{</think>\textbackslash n\textbackslash n}) and wraps the thinking content with a special token (e.g., <reason>). The specific token choice is not essential; we evaluate three variants: <reason>, \texttt{<begin>}, and \texttt{<less think>}.}
\label{tab:prompt_formats_oneline}
\end{table*}

\subsection{"Overfitting" in Reasoning Initiation}


Mainstream reasoning models, such as DeepSeek~\cite{liu2024deepseek,guo2025deepseek} and Qwen3~\cite{yang2025qwen3} thinking models, typically enclose their reasoning processes using explicit <think> tags. During the reasoning phase, these models often follow a fixed lexical pattern to initiate their thoughts, commonly starting with tokens such as ``Okay'' or ``Alright''. 

Beyond model outputs, a substantial portion of existing fine-tuning data is derived from generations produced by these reasoning models themselves. As a result, such datasets are heavily populated with these unified reasoning patterns and are subsequently used to fine-tune new models. Representative examples include OpenR1-Math-220k~\cite{openr1}, OpenMathReasoning~\citep{moshkov2025aimo2} and so on, which is summarized in \Cref{tab:token-model-three}.


\subsection{Attention-Based Evidence of Token-Level Triggers}
A natural question is whether such unified opening patterns may lead to "overfitting", where a small set of tokens dominates the model reasoning behavior and effectively serves as a switch for different reasoning modes. 

To investigate this question, we take Qwen3-8B as a case study and examine five different generation settings: (1) No-Think mode, (2) standard Think mode, (3) a generation starting with the pattern </think> followed by ``Okay'', (4) a think-style generation without explicit <think> tags and (5) a generation starting with the pattern <reason> followed by ``Okay''. For each setting, we let the model generate tokens and compute the average attention from subsequent tokens to the opening tokens. Specifically, we average attention weights across all layers and attention heads. 

The results are shown in \Cref{fig:attention_modes}. We find that in the 4 settings: Think mode, No <think>+``Okay'', </think>+``Okay'' and  <reason>+``Okay'', the token ``Okay'' consistently receives the highest attention from later tokens. This suggests that the model’s reasoning behavior is driven primarily by the lexical cue ``Okay''. By contrast, in the No-Think mode, attention is predominantly concentrated on the \textbackslash n\textbackslash n followed by </think> token. 

Together, these observations indicate that the \textit{model uses ``Okay'' as a trigger to enter the reasoning mode, while relying on the \textbackslash n\textbackslash n followed by </think> token as the primary signal to activate the No-Think behavior}. To further quantify this phenomenon, we compute the trigger-token attention mass for each reasoning mode and report detailed results in \Cref{sec:quant_attn}.

\subsection{Experimental Verification of Token-Level Triggers}






To validate the hypothesized ``overfitting'' of reasoning models to token-level cues, we construct compositional prompting modes combining the triggers discussed above: No-think, Think, No-think + Okay, No \texttt{<think>} + Okay, and \texttt{<reason>} + Okay (see \Cref{tab:format_mode_dataset} for exact formats).

We evaluate Qwen3-8B on MATH500, AIME22--24, and GPQA, reporting accuracy and wait count in \Cref{tab:format_mode_dataset}. Modes containing ``Okay'' as the opening cue (No \texttt{<think>} + Okay, \texttt{<reason>} + Okay) consistently match Think-mode accuracy and wait count, confirming its role as a reasoning trigger. The No-think + Okay setting, however, shows degraded performance, as the co-occurrence of \texttt{\textbackslash n\textbackslash n</think>} and ``Okay'' induces conflicting signals, yielding an intermediate rather than a clean reasoning regime.

\section{Mid-Think: Intermediate-Budget Reasoning without Training}

\subsection{Mid-Think: Implementation}

Building on the observations from the previous section, we find that in the Think mode, generated tokens primarily attend to the ``Okay'' cue following the <think> tag, whereas in the No-think mode, attention concentrates on the newline pattern following the </think> tag. Leveraging this insight, we combine these two cues to induce an intermediate reasoning regime without training, enabling the model to reason with a reduced token budget.

As shown in \Cref{tab:prompt_formats_oneline}, we introduce a new prompting format, termed Mid-Think. Concretely, the format is: <think>\textbackslash n\textbackslash n</think>\textbackslash n\textbackslash n <reason>\textbackslash nOkay\ldots. This design exploits the joint presence of the </think> newline cue and the trigger ``Okay'', allowing the model to be simultaneously influenced by reasoning and non-reasoning signals.

The <reason> tag is introduced to explicitly delimit the reasoning content, serving a structural role analogous to that of the <think> tag. Notably, the choice of <reason> is not essential; alternative tokens can be used to achieve the same effect. In our experiments, we consider three variants: <reason>, <begin>, and <less think>.

\begin{figure}[t]
    \centering
     \includegraphics[width=\linewidth]{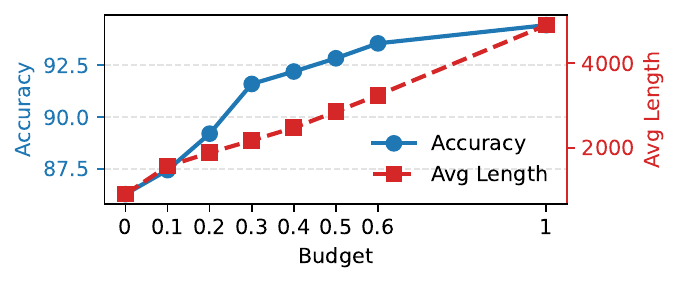}
     \vspace{-24pt}
    \caption{
        Budget-controlled reasoning on MATH500 using Qwen3-14B. The figure reports the average length and accuracy under different reasoning budgets. Both metrics increase steadily as the budget grows, validating the effectiveness of the budget-control mechanism.
        }
    \label{fig:verify different_budget}
\end{figure}

\subsection{Mid-Think Achieves Pareto-Optimal Performance}


To validate the effectiveness of Mid-Think, we first introduce a budget-controlled evaluation protocol that measures a reasoning model’s performance under different budgets in the standard Think mode, enabling us to construct a budget--performance curve. We then place Mid-Think on this curve and observe that it achieves intermediate-budget reasoning behavior and, in some cases, attains Pareto-optimal performance relative to similar budgets.


\begin{figure}[t]
    \centering
    \includegraphics[width=\linewidth]{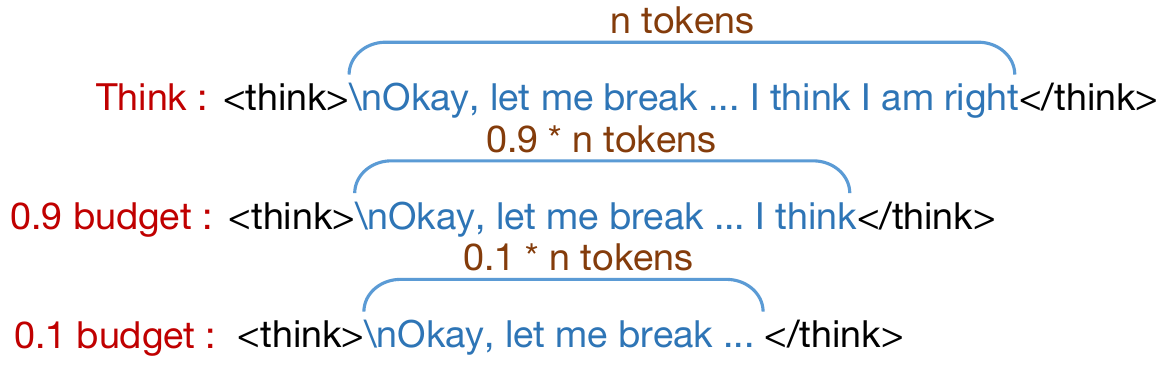}
     \vspace{-20pt}
    \caption{Overview of the budget-controlled method. The model first generates a full response. The reasoning (think) content is then truncated to the specified budget (in tokens) and concatenated with the remaining prompt, after which the model generates the final response.}
    \label{fig:different_budget}
\end{figure}

\subsubsection{Budget-Controlled Reasoning Baseline}

\begin{figure*}[t]
    \centering

    \begin{subfigure}[t]{0.32\textwidth}
        \centering
        \includegraphics[width=\linewidth]{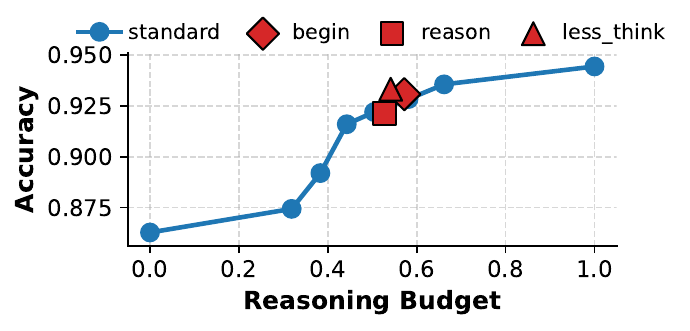}
        \caption{Qwen3-14B on MATH500}
    \end{subfigure}
    \hfill
    \begin{subfigure}[t]{0.32\textwidth}
        \centering
        \includegraphics[width=\linewidth]{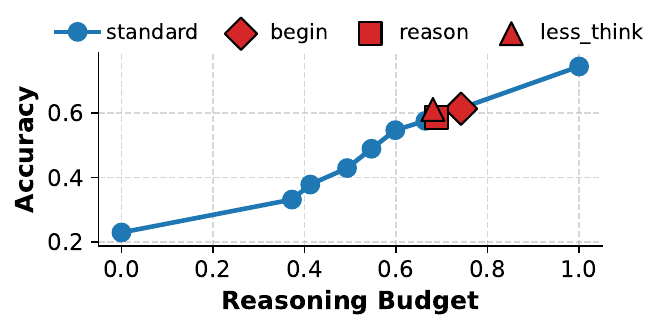}
        \caption{Qwen3-14B on AIME}
    \end{subfigure}
    \hfill
    \begin{subfigure}[t]{0.32\textwidth}
        \centering
        \includegraphics[width=\linewidth]{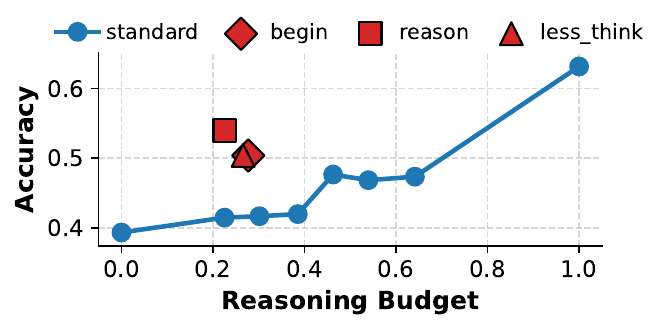}
        \caption{Qwen3-14B on GPQA}
    \end{subfigure}

    \vspace{0.8em}

    \begin{subfigure}[t]{0.32\textwidth}
        \centering
        \includegraphics[width=\linewidth]{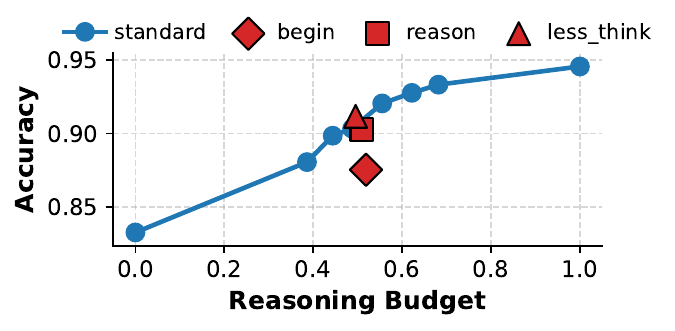}
        \caption{Qwen3-8B on MATH500}
    \end{subfigure}
    \hfill
    \begin{subfigure}[t]{0.32\textwidth}
        \centering
        \includegraphics[width=\linewidth]{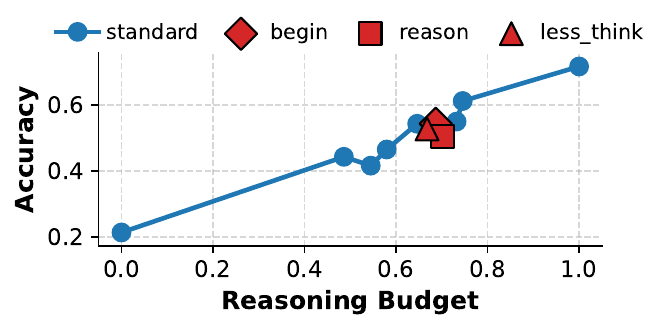}
        \caption{Qwen3-8B on AIME}
    \end{subfigure}
    \hfill
    \begin{subfigure}[t]{0.32\textwidth}
        \centering
        \includegraphics[width=\linewidth]{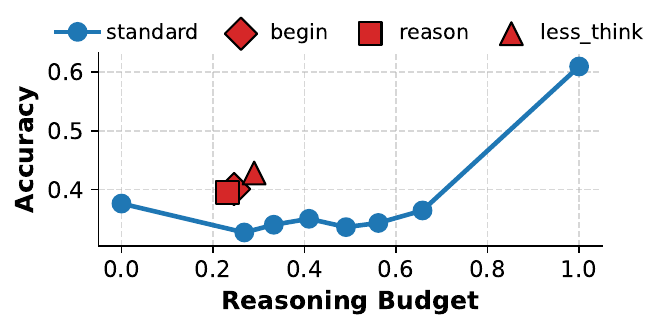}
        \caption{Qwen3-8B on GPQA}
    \end{subfigure}

    \vspace{0.8em}

    \begin{subfigure}[t]{0.32\textwidth}
        \centering
        \includegraphics[width=\linewidth]{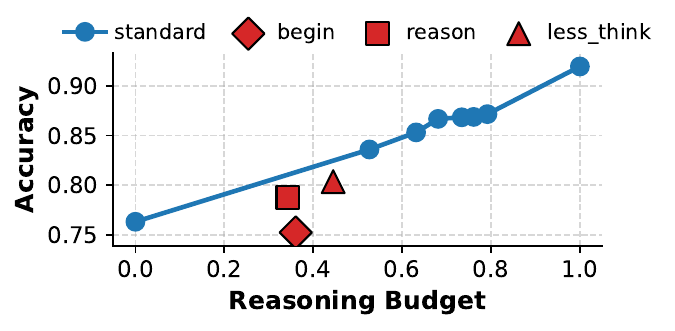}
        \caption{DeepSeek-7B on MATH500}
    \end{subfigure}
    \hfill
    \begin{subfigure}[t]{0.32\textwidth}
        \centering
        \includegraphics[width=\linewidth]{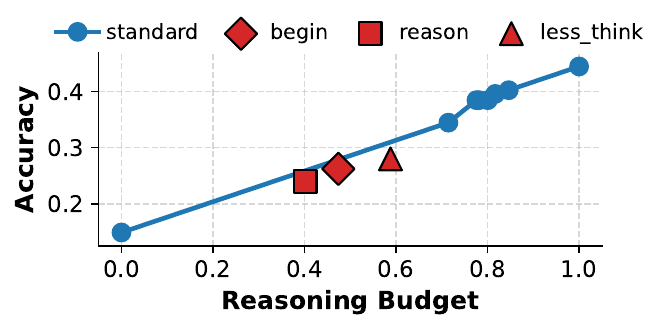}
        \caption{DeepSeek-7B on AIME}
    \end{subfigure}
    \hfill
    \begin{subfigure}[t]{0.32\textwidth}
        \centering
        \includegraphics[width=\linewidth]{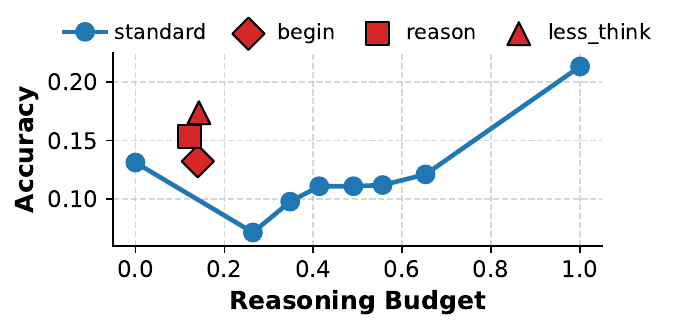}
        \caption{DeepSeek-7B on GPQA}
    \end{subfigure}

    \vspace{0.8em}

    \begin{subfigure}[t]{0.32\textwidth}
        \centering
        \includegraphics[width=\linewidth]{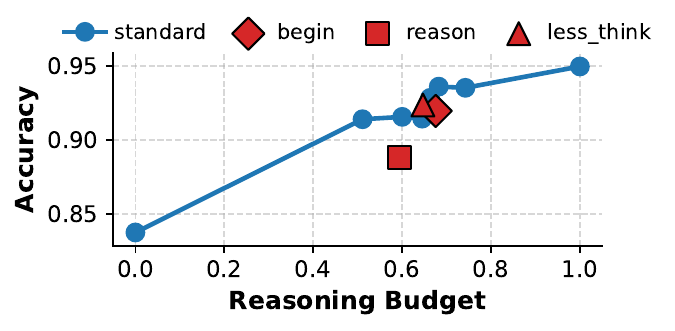}
        \caption{Qwen3-32B on MATH500}
    \end{subfigure}
    \hfill
    \begin{subfigure}[t]{0.32\textwidth}
        \centering
        \includegraphics[width=\linewidth]{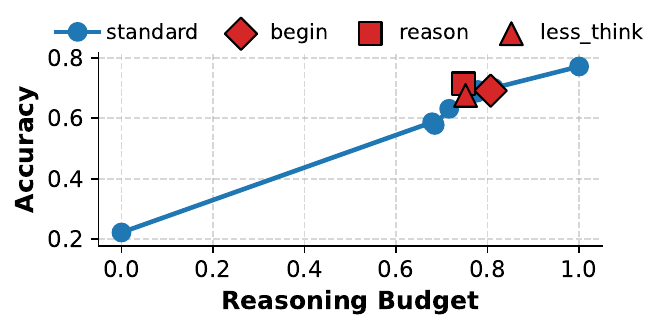}
        \caption{Qwen3-32B on AIME}
    \end{subfigure}
    \hfill
    \begin{subfigure}[t]{0.32\textwidth}
        \centering
        \includegraphics[width=\linewidth]{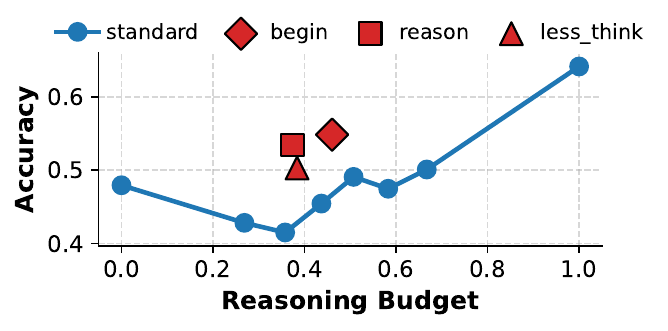}
        \caption{Qwen3-32B on GPQA}
    \end{subfigure}

    \caption{
        Comparison between models under different reasoning budgets and Mid-Think across multiple datasets and model types, including hybrid-thinking (Qwen3-8B, Qwen3-14B, Qwen3-32B) and pure-think (DeepSeek-R1-Distill-Qwen-7B) models. Mid-Think is evaluated with different tags (\texttt{<reason>}, \texttt{<begin>}, \texttt{<less think>}). Across all models and datasets (MATH500, AIME, and GPQA), Mid-Think consistently achieves performance corresponding to intermediate reasoning budgets, and on GPQA it even surpasses fixed-budget baselines, yielding Pareto-optimal accuracy--efficiency trade-offs between Think and No-Think mode.
    }
    \label{fig:token_cue_all}
\end{figure*}

\textbf{Implementation.} To obtain a model’s reasoning capability under different budgets, we first let the reasoning model generate a complete reasoning trajectory in the standard Think mode. Suppose the full reasoning process contains $n$ tokens. We then construct budget-controlled variants by retaining only a fraction of the reasoning tokens according to a predefined budget ratio.

Concretely, under a budget of $0.9$, we keep the first $0.9\times n$ reasoning tokens, whereas under a budget of $0.1$, only the first $0.1\times n$ tokens are preserved, as illustrated in \Cref{fig:different_budget}. This procedure allows us to systematically control the effective reasoning budget and obtain reasoning performance across different budget levels.

\textbf{Verification.} To verify the effectiveness, we conduct experiments using Qwen3-14B on MATH500. We evaluate the model under multiple budget settings, including $0.1, 0.2, 0.3, 0.4, 0.5, 0.6$, as well as the No-Think setting (corresponding to a budget of $0.0$) and the standard Think setting (corresponding to a budget of $1.0$).

As shown in \Cref{fig:verify different_budget}, both the average generation length and accuracy increase monotonically with the reasoning budget. This behavior is consistent with the intended effect of the budget-control mechanism, confirming that the proposed method provides a reliable and interpretable way to modulate the model’s effective reasoning capacity.

\begin{table*}[h]
\centering
\small
\setlength{\tabcolsep}{4pt}
\begin{tabular}{lcccccc}
\toprule
\multirow{2}{*}{\textbf{Model}} & \multicolumn{3}{c}{\textbf{Acc on MATH-500}} & \multicolumn{3}{c}{\textbf{Avg Length}} \\
\cmidrule(lr){2-4} \cmidrule(lr){5-7}
& No-T & Mid-T & Think & No-T & Mid-T & Think \\
\midrule
Phi-4-mini-reasoning         & 75.6 & \textbf{89.4} & 90.6 & 1{,}047 & 3{,}333 & 4{,}031 \\
DeepSeek-R1-Distill-Llama-8B & 60.8 & \textbf{72.6} & 85.6 & 1{,}045 & 2{,}062 & 4{,}345 \\
DeepSeek-R1-Distill-Qwen-2.5-7B & 68.2 & \textbf{74.6} & 86.8 & 718 & 1{,}655 & 3{,}957 \\
Qwen3-30B-A3B (MoE)          & 87.2 & \textbf{92.0} & 94.0 & 898 & 3{,}007 & 5{,}099 \\
\bottomrule
\end{tabular}
\caption{Accuracy and average generation length on MATH-500 across different model families and architectures. Mid-Think (Mid-T) consistently improves over No-Think (No-T) while using fewer tokens than Think mode.}
\label{tab:cross_model}
\end{table*}
\begin{table*}[h]
\centering
\small
\begin{tabular}{lcccccc}
\toprule
\multirow{2}{*}{\textbf{Model}} & \multicolumn{3}{c}{\textbf{Acc on LiveCodeBench}} & \multicolumn{3}{c}{\textbf{Avg Length}} \\
\cmidrule(lr){2-4} \cmidrule(lr){5-7}
& No-T & Mid-T & Think & No-T & Mid-T & Think \\
\midrule
Qwen3-8B  & 40.8 & \textbf{49.7} & 65.9 & 760 & 2{,}212 & 9{,}931 \\
Qwen3-14B & 46.8 & \textbf{62.7} & 72.3 & 509 & 2{,}147 & 8{,}959 \\
\bottomrule
\end{tabular}
\caption{Accuracy and average generation length on LiveCodeBench. Mid-Think (Mid-T) generalizes beyond math reasoning tasks, consistently outperforming No-Think (No-T) with significantly shorter outputs than Think mode.}
\label{tab:livecode}
\end{table*}

\subsubsection{Mid-Think v.s. Different Budget }



\textbf{Experimental Setup.} We evaluate a diverse set of models, including hybrid reasoning models (Qwen3-8B and Qwen3-14B), a pure supervised model (DeepSeek-Qwen-7B), and a purely RL-trained model (Qwen3-32B). For each model, we first measure performance under different fixed budgets ranging from $0.1$ to $0.6$, as well as the standard Think and No-think settings, on MATH500, AIME22--24, and GPQA. We then evaluate the same models under the proposed Mid-Think mode and visualize all results jointly for comparison.

\textbf{Hybrid-Think Model Results.} \Cref{fig:token_cue_all} presents the results of Qwen3-8B and Qwen3-14B on MATH500, AIME22--24, and GPQA. Across all three benchmarks, the performance of the proposed Mid-Think mode consistently aligns with that of an intermediate budget, approximately corresponding to a budget of $0.5$ in terms of accuracy.

Notably, on GPQA, Mid-Think achieves the strongest performance, surpassing neighboring budget settings and even exceeding the Pareto frontier defined by the budget--performance trade-off. These results indicate that reasoning models with Mid-Think mode can obtain intermediate-budget reasoning capability in a training-free manner.

Similar trends are observed for Qwen3-14B, demonstrating that the effectiveness of Mid-Think generalizes across different model scales.

\textbf{Pure-Think Model Results.} To examine the effectiveness of Mid-Think on pure reasoning models, we additionally evaluate DeepSeek-Qwen-7B, with results shown in \Cref{fig:token_cue_all}. Under Mid-Think setting, the model’s accuracy tends to align with smaller-budget regimes. We attribute this behavior to the model placing stronger emphasis on the </think>\textbackslash n\textbackslash n pattern, which biases the effective reasoning budget toward shorter reasoning spans.

Nevertheless, on GPQA, Mid-Think still achieves improved performance, surpassing the Pareto frontier defined by near-budget. This result suggests that even for pure thinking models, Mid-Think can yield favorable trade-offs between reasoning budget and performance.

\textbf{RL-Model Results.} To assess the effectiveness on RL-trained models, we evaluate Qwen3-32B on MATH500, AIME22--24, and GPQA, with results shown in \Cref{fig:token_cue_all}. We observe that RL-trained model remains compatible with the Mid-Think mode, exhibiting performance that closely matches an intermediate budget of approximately $0.5$.

Moreover, on GPQA, Mid-Think again surpasses the Pareto frontier defined by near-budget. These findings indicate that the proposed Mid-Think strategy generalizes to RL-trained models and enables favorable budget--performance trade-offs even at larger model scales.

\subsection{Generalization Across Model Families and Architectures}

\paragraph{Across Model Families and Architectures.}
To assess whether Mid-Think generalizes beyond the Qwen family, we evaluate on four additional reasoning models spanning different families and architectures: Phi-4-mini-reasoning (Phi, dense), DeepSeek-R1-Distill-Llama-8B (Llama, dense), DeepSeek-R1-Distill-Qwen-2.5-7B (Qwen, dense), and Qwen3-30B-A3B (Qwen, MoE).
All models were trained on reasoning traces beginning with ``\texttt{Okay}''.
As shown in \Cref{tab:cross_model}, Mid-Think consistently outperforms No-Think across all models and architectures while remaining substantially shorter than Think mode, demonstrating that the method is model-family and architecture agnostic.

\paragraph{Beyond Math: Coding Tasks.}
To assess whether Mid-Think generalizes beyond mathematical reasoning, we evaluate on LiveCodeBench\cite{jain2024livecodebench} using Qwen3-8B and Qwen3-14B.
As shown in \Cref{tab:livecode}, Mid-Think maintains its effectiveness on this coding benchmark: it substantially outperforms No-Think in accuracy while generating far fewer tokens than full Think mode.
This demonstrates that the token-level trigger mechanism is not specific to mathematical reasoning but generalizes across task domains.

\subsection{Trigger Origin and Robustness}
A natural follow-up question is whether the observed trigger dependence is an emergent property of model \emph{architecture}, or a consequence of \emph{training data templates}.
Our experiments support the latter. Across Qwen, Llama, and Phi model families---all of which were trained on reasoning traces beginning with ``\texttt{Okay}''. To further isolate the effect of training templates, we test \texttt{Alibaba-Apsara/DASD-4B-Thinking}, a model whose reasoning traces begin with ``\texttt{We need}'' rather than ``\texttt{Okay}''.
Attention analysis on DASD-4B confirms that the dominant trigger token shifts from ``\texttt{Okay}'' to ``\texttt{We}'', suggesting the trigger adapts to the training template.
Replacing the Mid-Think trigger accordingly (``\texttt{Okay}'' $\to$ ``\texttt{We}'') preserves Mid-Think's effectiveness, as shown in \Cref{tab:dasd_trigger}.

\begin{table}[t]
\centering
\small
\begin{tabular}{lccc}
\toprule
\textbf{Mode} & \textbf{Trigger} & \textbf{Acc (\%)} & \textbf{Avg Len} \\
\midrule
No-Think  & ---              & 82.4 & 2{,}219 \\
Mid-Think & \texttt{We}      & 88.8 & 2{,}796 \\
Think     & ---              & 90.8 & 3{,}516 \\
\bottomrule
\end{tabular}
\caption{Mid-Think with a template-adapted trigger (``\texttt{We}'') on DASD-4B-Thinking (MATH-500). The method remains effective when the trigger is matched to the model's training template.}
\label{tab:dasd_trigger}
\end{table}

In contrast, applying a mismatched trigger (\emph{e.g.}, using ``\texttt{We}'' on Qwen3-8B, which was trained with ``\texttt{Okay}'') causes Mid-Think to collapse to near No-Think behavior, as shown in \Cref{tab:trigger_sensitivity}.
This confirms that the trigger must match the model's training template to be effective, and that the phenomenon originates from training data rather than model architecture.

\begin{table}[t]
\centering
\small
\begin{tabular}{lcc}
\toprule
\textbf{Mode} & \textbf{Acc (\%)} & \textbf{Avg Len} \\
\midrule
No-Think                       & 83.2 & 1{,}013 \\
Mid-Think (\texttt{We})        & 82.2 & 1{,}104 \\
Mid-Think (\texttt{Okay})      & 92.3 & 2{,}753 \\
Think                          & 94.6 & 5{,}557 \\
\bottomrule
\end{tabular}
\caption{Sensitivity to trigger token choice on Qwen3-8B (MATH-500). A mismatched trigger causes Mid-Think to degenerate to No-Think behavior, confirming trigger-template alignment is essential.}
\label{tab:trigger_sensitivity}
\end{table}

\subsection{Comparison with Fixed-Token and Prompt-Based Methods}

This section compares our approach with existing training-free methods for achieving intermediate reasoning capability. Most prior approaches rely on additional training to control the reasoning budget. In contrast, the primary training-free alternatives are the fixed-token budget mechanism in the Qwen3 series and prompt-based budget constraints. We therefore compare Mid-Think against these two training-free baselines.

\textbf{Experimental Setup}. To compare against existing training-free budget control methods, we use Qwen3-14B as the backbone model and evaluate three baselines on MATH500. First, we adopt the fixed-token budget strategy provided by the Qwen3 series, setting the maximum generation length to 2k, 3k, and 4k tokens, respectively. Second, we evaluate the proposed Mid-Think mode under the same evaluation setting. Finally, we test prompt-based budget control by applying prompting templates designed to constrain or guide the reasoning process. The results are summarized in \Cref{tab:budget_comparison_math500}.

\begin{table}[t]
\centering
\small
\setlength{\tabcolsep}{8pt}
\begin{tabular}{c c c c}
\toprule
\textbf{Method} & \textbf{Setting} & \textbf{Acc(\%)} & \textbf{Avg Len} \\
\midrule
 
\multirow{2}{*}{Original}
& No-Think & 86.3 & 899 \\
& Think & 94.4 & 4904 \\
\midrule
\multirow{3}{*}{Fixed Tokens}
& 2k tokens & 89.6 & 2673 \\
& 3k tokens & 91.2 & 3315 \\
& 4k tokens & 90.8 & 3793 \\
& 5k tokens & 92.2 & 4136 \\
\midrule
\textbf{Mid-Think}
& training-free & \textbf{92.1} & \textbf{2589} \\
\midrule
Prompt-based
& Prompt & 91.2 & 3131 \\
\bottomrule
\end{tabular}
\caption{
Comparison of training-free budget control methods on MATH500 using Qwen3-14B. Fixed Tokens limits the number of tokens allocated to the reasoning process, while Prompt-based methods prepend explicit instructions encouraging shorter outputs while preserving accuracy.
}
\label{tab:budget_comparison_math500}
\end{table}


\textbf{Comparison Results}. As shown in \Cref{tab:budget_comparison_math500}, while the Fixed Tokens strategy enables budget control by explicitly limiting the number of reasoning tokens, its control is relatively coarse. Under comparable accuracy levels, Mid-Think consistently achieves substantially shorter average generation lengths. Moreover, fixed tokens require pre-specifying the token limit without knowing the difficulty of each instance in advance, which often leads to insufficient reasoning for harder problems and consequently degrades accuracy, despite successfully constraining output length.

Prompt-based methods also underperform compared to Mid-Think. Although such prompts can partially reduce the reasoning budget, the resulting average generation length remains significantly higher than that achieved by Mid-Think, indicating limited controllability. In contrast, Mid-Think leverages overfitted token-level triggers to induce intermediate-budget reasoning behavior, providing a more reliable and fine-grained mechanism for balancing reasoning quality and output length.

\section{Applying Mid-Think to RL Training after SFT}


This section applies the proposed Mid-Think mode to RL training on top of supervised fine-tuning (SFT). Prior work has explored RL training using No-think settings~\cite{xu2025thinking} or fixed-token budgets~\cite{xu2025scalable} to accelerate and improve reasoning training. We show that the Mid-Think mode can also be directly trained via RL, enabling more efficient training while improving overall model performance.

\subsection{Experimental Setup}


In this section, we conduct RL training using Qwen3-4B and Qwen3-8B. All models are trained with the \texttt{verl} framework using the GRPO algorithm, with a learning rate of $1\times10^{-6}$ for six training epochs. We set the maximum generation length to 16K tokens during training. All experiments are conducted using 8 NVIDIA H200 GPUs.

We use 2 training datasets, each consisting of 5{,}000 samples, sampled from Skywork-OR1-RL-Data~\cite{he2025skywork,skywork-or1-2025} and OpenScienceReasoning-2~\cite{nvidia2025nvidianemotronnano2}. Models trained on Skywork-OR1-RL-Data are evaluated on AIME, while models trained on OpenScienceReasoning-2 are evaluated on GPQA. This evaluation enables validation across both mathematical and scientific reasoning domains.

\begin{table*}[t]
\centering
\small
\setlength{\tabcolsep}{3.5pt}
\begin{tabular}{c c c c c c c c c |c c c c c c}
\toprule
\multirow{3}{*}{\textbf{Model}} &
\multirow{3}{*}{\textbf{Training}} &
\multirow{3}{*}{\textbf{Mode}} &
\multicolumn{6}{c}{\textbf{AIME}} &
\multicolumn{6}{c}{\textbf{GPQA}} \\
\cmidrule(lr){4-9} \cmidrule(lr){10-15}
& & &
\multicolumn{3}{c}{\textbf{No-think Test}} &
\multicolumn{3}{c}{\textbf{Think Test}} &
\multicolumn{3}{c}{\textbf{No-think Test}} &
\multicolumn{3}{c}{\textbf{Think Test}} \\
\cmidrule(lr){4-6} \cmidrule(lr){7-9}
\cmidrule(lr){10-12} \cmidrule(lr){13-15}
& & &
Acc & Len & Wait &
Acc &  Len & Wait &
Acc &  Len & Wait &
Acc &  Len & Wait \\
\midrule
\multirow{4}{*}{Qwen3-8B}
& No & --
& \textcolor{gray}{21.3} & \textcolor{gray}{4520} & \textcolor{gray}{447}
& \textcolor{gray}{71.6} & \textcolor{gray}{16847} & \textcolor{gray}{40342}
& \textcolor{gray}{37.5} & \textcolor{gray}{1335} & \textcolor{gray}{459}
& \textcolor{gray}{60.9} & \textcolor{gray}{9008} & \textcolor{gray}{71817} \\
& RL & Think
& 19.3 & 7624 & 9423
& 69.8 & 13330 & 34701
& 34.7 & 1567 & 2179
& 58.5 & 9413 & 52228 \\
& RL & No-Think
& 62.9 & 13591 & 11936
& 72.0 & 18980 & 34585
& 47.8 & 4435 & 9863
& 59.6 & 9047 & 59419 \\
& RL & Mid-Think
& 27.6 & 7114 & 6055
& \textbf{72.4} & 15318 & 44142
& 36.2 & 1293 & 1968
& \textbf{61.1} & 8257 & 55272 \\
\midrule
\multirow{4}{*}{\texttt{Qwen3-4B}}
& No & --
& \textcolor{gray}{20.0} & \textcolor{gray}{4605} & \textcolor{gray}{542}
& \textcolor{gray}{68.7} & \textcolor{gray}{16699} & \textcolor{gray}{40089}
& \textcolor{gray}{36.9} & \textcolor{gray}{1560} & \textcolor{gray}{590}
& \textcolor{gray}{53.2} & \textcolor{gray}{8729} & \textcolor{gray}{72509} \\
& RL & Think
& 17.6 & 7973 & 917
& 61.1 & 13347 & 32118
& 35.5 & 1789 & 252
& 47.3 & 10481 & 58476 \\
& RL & No-Think
& 36.7 & 17421 & 20818
& 69.3 & 20551 & 61341
& 40.4 & 6871 & 17196
& 48.7 & 10774 & 82568 \\
& RL & Mid-Think
& 24.9 & 12492 & 11271
& \textbf{69.6} & 15799 & 40216
& 38.2 & 2086 & 4790
& \textbf{55.3} & 8696 & 60790 \\
\bottomrule
\end{tabular}
\vspace{-8pt}
\caption{
Performance of \texttt{Qwen3-4B} and Qwen3-8B after GRPO training under different modes (Think, No-think, and Mid-Think) on AIME and GPQA. We report accuracy, average generation length, and wait count. 
The Direct setting corresponds to the untrained model. 
No-think Test evaluates the model under the no-thinking mode, while Think Test evaluates the model under the standard thinking mode. Models trained with Mid-Think consistently outperform those trained with Think or No-think when evaluated in the Think mode, while largely preserving performance under the No-think  test setting.
}
\label{tab:test_mode_split_full}
\end{table*}


\begin{figure}[t]
    \centering
     \includegraphics[width=0.7\linewidth]{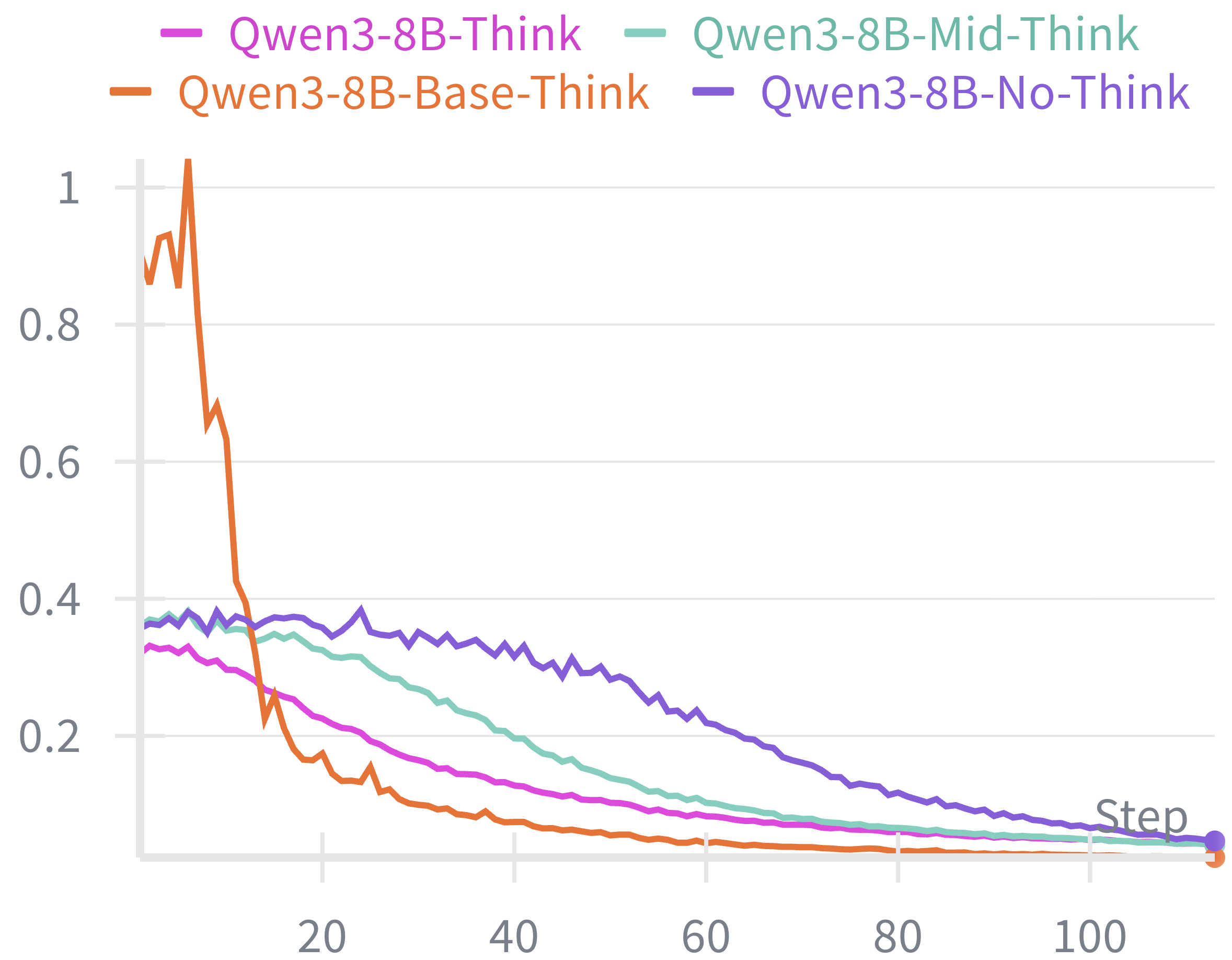}
     \vspace{-8pt}
    \caption{
        Entropy during GRPO. We compare Qwen3-8B-Base trained in the Think mode with Qwen3-8B trained under Think, No-think, and Mid-Think modes. The panel plots entropy versus training steps. Notably, Mid-Think both increases training entropy.
        }
    \label{fig:entropy}
\end{figure}

\begin{figure}[t]
    \centering
     \includegraphics[width=0.7\linewidth]{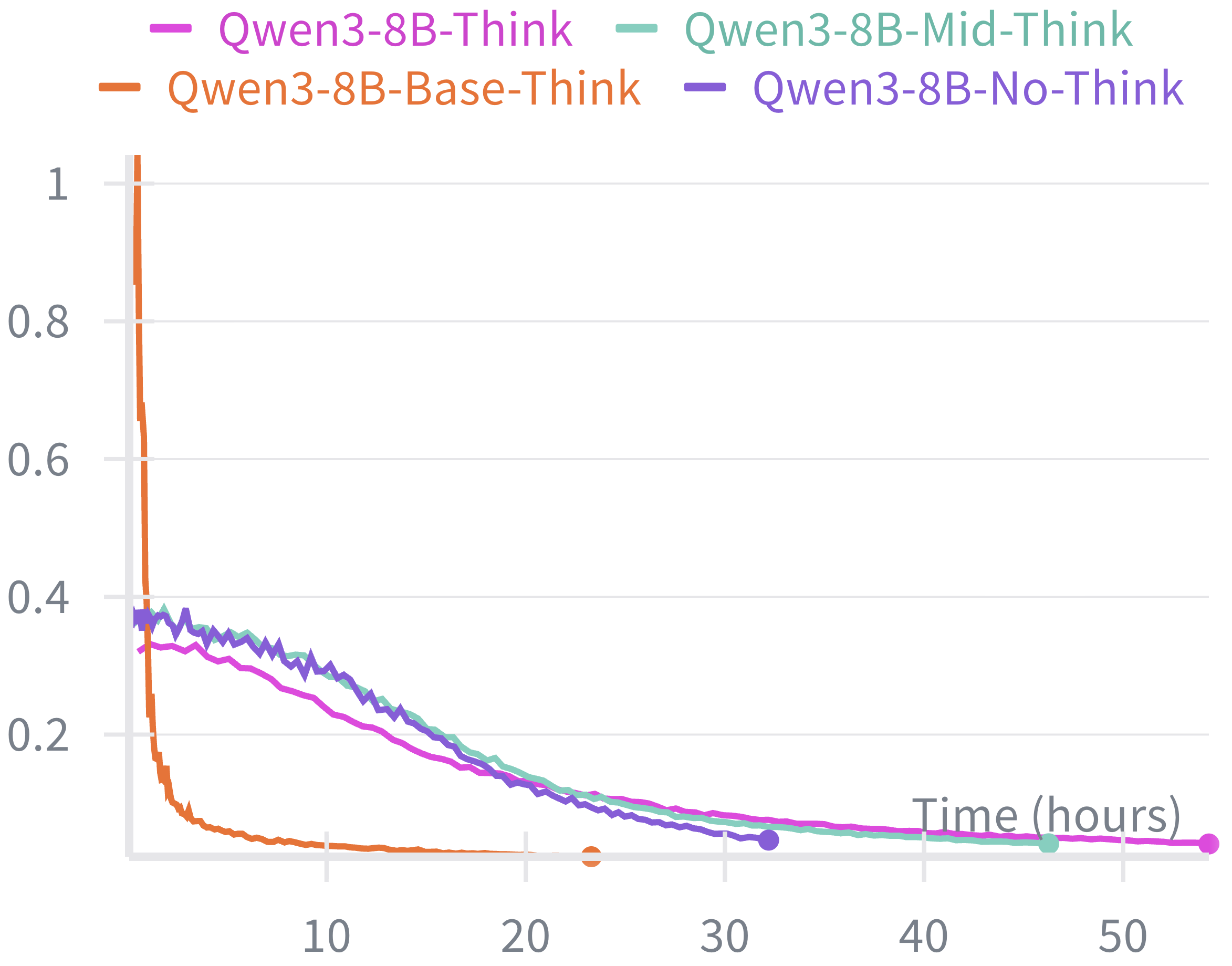}
     \vspace{-8pt}
    \caption{
        Entropy during GRPO. We compare Qwen3-8B-Base trained in the Think mode with Qwen3-8B trained under Think, No-think, and Mid-Think modes. The panel shows entropy versus relative training time. Notably, Mid-Think reduces overall training time.
        }
    \label{fig:entropy_time}
\end{figure}

\subsection{Experimental Results}

\subsubsection{Results of Training Entropy}
We analyze the evolution of training entropy for Qwen3-8B under different training modes, as illustrated in \Cref{fig:entropy}. Specifically, we report entropy curves on math training data for Qwen3-8B-Base trained in the Think mode, Qwen3-8B trained in the Think mode, and Qwen3-8B trained under the No-think and Mid-Think modes.

Qwen3-8B-Base trained in the Think mode exhibits the highest initial entropy, which rapidly decreases and eventually collapses. In contrast, Qwen3-8B trained in the Think mode starts with substantially lower entropy and remains in a low-entropy regime throughout training. We attribute this behavior to the fact that Qwen3-8B, after SFT, already exhibits a highly fixed reasoning pattern in the Think mode, thereby limiting exploration during RL training.

Notably, both the No-think and Mid-Think training modes lead to higher and more sustained entropy. This indicates that relaxing or partially disrupting the fixed reasoning pattern can effectively increase policy entropy, facilitating exploration and improving the stability of RL training.

\subsubsection{Results of Training Time}
\Cref{fig:entropy_time} reports the training time under different training modes. We observe that Qwen3-8B trained in the Think mode incurs long training time, as the model generates excessively long reasoning sequences during optimization. In contrast, Qwen3-8B-Base exhibits much shorter training time since it has not undergone prior fine-tuning and does not produce explicit reasoning traces.

Notably, both the No-think and Mid-Think modes significantly reduce training time by shortening the model’s reasoning outputs. This demonstrates that controlling the reasoning budget not only affects learning dynamics but also leads to substantial gains in training efficiency.

\subsubsection{Test Performance.}
We report the test performance of Qwen3-8B and Qwen3-4B after RL training under different modes on MATH500, AIME, and GPQA, including accuracy, average generation length, and wait count, as summarized in~\Cref{tab:test_mode_split_full}. 

Across all benchmarks, models trained in the standard Think mode consistently underperform those trained with the No-think and Mid-Think modes. Moreover, after No-think training, the model fails to reliably preserve non-reasoning behavior and tends to produce excessively long outputs when evaluated in the Think mode.

In contrast, RL training with the proposed Mid-Think mode achieves a more favorable balance. It largely maintains the efficiency benefits of the No-think regime while simultaneously attaining the highest accuracy under the Think mode.

\section{Related Works}

\textbf{Efficient LLM Reasoning.} Recent reasoning models still face significant efficiency challenges, often producing excessively long outputs~\citep{bandyopadhyay2025thinking, li2025system,wang2025demystifying}. Early approaches such as Kimi~1.5~\citep{team2025kimi} and Sky-Thought~\citep{reduce_overthinking_2025} reduce verbosity by aligning long and short responses via preference optimization, while TokenSkip~\citep{xia2025tokenskip} and LightThinker~\citep{zhang2025lightthinker} improve efficiency by pruning redundant tokens or compressing intermediate thoughts. Beyond shortening reasoning traces, hybrid thinking methods~\citep{jiang2025think,liu2025reinforcement} aim to control \emph{when} models reason, typically through explicit control tokens (e.g., \texttt{\textbackslash think}, \texttt{\textbackslash nothink})~\citep{sui2025stop, chen2024not,yang2026ace}. This paradigm has been adopted by models such as Gemini~\citep{team2025gemma}, Qwen3~\citep{yang2025qwen3}, GPT-oss~\citep{agarwal2025gpt,zhang2025beyond,yue2025don}, and DeepSeek~V3.1~\citep{liu2024deepseek}, with the latter further scaling hybrid thinking through large-scale RL.

\textbf{Reinforcement Learning with Verifiable Rewards}
With the emergence of DeepSeek-R1~\cite{guo2025deepseek,liu2024deepseek}, GRPO has become a widely adopted approach for endowing language models with reasoning capabilities~\cite{zhang2025survey,plaat2024reasoning,xu2025towards,yang2025longer}. A growing body of work focuses on improving GRPO, including variants~\cite{xi2025bapo,nan2025ngrpo,yangspeculative} such as DAPO~\cite{yu2025dapo}, Dr.\ GRPO~\cite{liu2025understanding}, and GSPO~\cite{zheng2025group}. Other studies specifically address the issue of entropy collapse in GRPO-based training like \emph{Rethinking Entropy Interventions}~\cite{hao2025rethinking,ganguly2026trust}. Several works aim to reduce the high training cost of GRPO, like \emph{It Takes Two}~\cite{wu2025takes},  \emph{Thinking-Free Policy}. In parallel, recent research has examined the interaction between supervised fine-tuning (SFT) and GRPO-based RL
 like \emph{On the Interplay}~\cite{zhang2025interplay} and \emph{Quagmires in SFT--RL Post-Training}~\cite{kang2025quagmires,yang2025100}. 


\section{Conclusion}

We show that hybrid thinking is largely driven by a small set of token-level triggers. Building on this, Mid-Think enables training-free intermediate-budget reasoning by strategically manipulating these triggers at inference time, achieving a better accuracy--efficiency trade-off than fixed-budget baselines. Beyond inference-time control, Mid-Think also proves effective as an RL training objective, yielding models that better balance Think and No-Think behavior. We hope these findings offer a lightweight and practical lens for understanding reasoning in hybrid thinking models.

\newpage
\section*{Limitations}


While Mid-Think enables intermediate-budget reasoning, it does not provide fully dynamic or fine-grained control over arbitrary budget levels (e.g., 0.1 or 0.7). Moreover, Mid-Think relies on the presence of existing overfitted token-level behaviors, and thus requires identifying such patterns in advance to realize Mid-Think.

\section*{Acknowledgements}
This work was supported in part by NSF award 2117439. This research made use of the High Performance Computing Resource in the Core Facility for Advanced Research Computing at Case Western Reserve University (CWRU).


\bibliography{custom}

@misc{openr1,
    title = {Open R1: A fully open reproduction of DeepSeek-R1},
    url = {https://github.com/huggingface/open-r1},
    author = {{Hugging Face}},
    month = {January},
    year = {2025}
}

@inproceedings{rein2024gpqa,
  title={Gpqa: A graduate-level google-proof q\&a benchmark},
  author={Rein, David and Hou, Betty Li and Stickland, Asa Cooper and Petty, Jackson and Pang, Richard Yuanzhe and Dirani, Julien and Michael, Julian and Bowman, Samuel R},
  booktitle={First Conference on Language Modeling},
  year={2024}
}

@article{liu2025understanding,
  title={Understanding R1-Zero-Like Training: A Critical Perspective},
  author={Zichen Liu and Changyu Chen and Wenjun Li and Penghui Qi and Tianyu Pang and Chao Du and Wee Sun Lee and Min Lin},
  journal={arXiv preprint arXiv:2503.20783},
  year={2025}
}

@article{yu2025dapo,
  title={DAPO: An Open-Source LLM Reinforcement Learning System at Scale},
  author={Yu, Qiying and Zhang, Zheng and Zhu, Ruofei and Yuan, Yufeng and Zuo, Xiaochen and Yue, Yu and Fan, Tiantian and Liu, Gaohong and Liu, Lingjun and Liu, Xin and others},
  journal={arXiv preprint arXiv:2503.14476},
  year={2025}
}

@article{li2025system,
  title={From system 1 to system 2: A survey of reasoning large language models},
  author={Li, Zhong-Zhi and Zhang, Duzhen and Zhang, Ming-Liang and Zhang, Jiaxin and Liu, Zengyan and Yao, Yuxuan and Xu, Haotian and Zheng, Junhao and Wang, Pei-Jie and Chen, Xiuyi and others},
  journal={arXiv preprint arXiv:2502.17419},
  year={2025}
}

@article{bandyopadhyay2025thinking,
  title={Thinking Machines: A Survey of LLM based Reasoning Strategies},
  author={Bandyopadhyay, Dibyanayan and Bhattacharjee, Soham and Ekbal, Asif},
  journal={arXiv preprint arXiv:2503.10814},
  year={2025}
}

@article{plaat2024reasoning,
  title={Reasoning with large language models, a survey},
  author={Plaat, Aske and Wong, Annie and Verberne, Suzan and Broekens, Joost and van Stein, Niki and Back, Thomas},
  journal={arXiv preprint arXiv:2407.11511},
  year={2024}
}

@article{team2025kimi,
  title={Kimi k1. 5: Scaling reinforcement learning with llms},
  author={Team, Kimi and Du, Angang and Gao, Bofei and Xing, Bowei and Jiang, Changjiu and Chen, Cheng and Li, Cheng and Xiao, Chenjun and Du, Chenzhuang and Liao, Chonghua and others},
  journal={arXiv preprint arXiv:2501.12599},
  year={2025}
}

@article{liu2024deepseek,
  title={Deepseek-v2: A strong, economical, and efficient mixture-of-experts language model},
  author={Liu, Aixin and Feng, Bei and Wang, Bin and Wang, Bingxuan and Liu, Bo and Zhao, Chenggang and Dengr, Chengqi and Ruan, Chong and Dai, Damai and Guo, Daya and others},
  journal={arXiv preprint arXiv:2405.04434},
  year={2024}
}

@article{guo2025deepseek,
  title={Deepseek-r1: Incentivizing reasoning capability in llms via reinforcement learning},
  author={Guo, Daya and Yang, Dejian and Zhang, Haowei and Song, Junxiao and Zhang, Ruoyu and Xu, Runxin and Zhu, Qihao and Ma, Shirong and Wang, Peiyi and Bi, Xiao and others},
  journal={arXiv preprint arXiv:2501.12948},
  year={2025}
}

@article{sui2025stop,
  title={Stop Overthinking: A Survey on Efficient Reasoning for Large Language Models},
  author={Sui, Yang and Chuang, Yu-Neng and Wang, Guanchu and Zhang, Jiamu and Zhang, Tianyi and Yuan, Jiayi and Liu, Hongyi and Wen, Andrew and Chen, Hanjie and Hu, Xia and others},
  journal={arXiv preprint arXiv:2503.16419},
  year={2025}
}

@misc{reduce_overthinking_2025,
  author       = {NovaSky Team},
  title        = {Think Less, Achieve More: Cut Reasoning Costs by 50% Without Sacrificing Accuracy},
  howpublished = {https://novasky-ai.github.io/posts/reduce-overthinking},
  note         = {Accessed: 2025-01-23},
  year         = {2025}
}

@misc{xia2025tokenskip,
      title={TokenSkip: Controllable Chain-of-Thought Compression in LLMs}, 
      author={Heming Xia and Yongqi Li and Chak Tou Leong and Wenjie Wang and Wenjie Li},
      year={2025},
      eprint={2502.12067},
      archivePrefix={arXiv},
      primaryClass={cs.CL},
      url={https://arxiv.org/abs/2502.12067}, 
}

@article{zhang2025lightthinker,
  title={Lightthinker: Thinking step-by-step compression},
  author={Zhang, Jintian and Zhu, Yuqi and Sun, Mengshu and Luo, Yujie and Qiao, Shuofei and Du, Lun and Zheng, Da and Chen, Huajun and Zhang, Ningyu},
  journal={arXiv preprint arXiv:2502.15589},
  year={2025}
}

@article{chen2024not,
  title={Do not think that much for 2+ 3=? on the overthinking of o1-like llms},
  author={Chen, Xingyu and Xu, Jiahao and Liang, Tian and He, Zhiwei and Pang, Jianhui and Yu, Dian and Song, Linfeng and Liu, Qiuzhi and Zhou, Mengfei and Zhang, Zhuosheng and others},
  journal={arXiv preprint arXiv:2412.21187},
  year={2024}
}

@article{agarwal2025gpt,
  title={gpt-oss-120b \& gpt-oss-20b model card},
  author={Agarwal, Sandhini and Ahmad, Lama and Ai, Jason and Altman, Sam and Applebaum, Andy and Arbus, Edwin and Arora, Rahul K and Bai, Yu and Baker, Bowen and Bao, Haiming and others},
  journal={arXiv preprint arXiv:2508.10925},
  year={2025}
}

@article{team2025gemma,
  title={Gemma 3 technical report},
  author={Team, Gemma and Kamath, Aishwarya and Ferret, Johan and Pathak, Shreya and Vieillard, Nino and Merhej, Ramona and Perrin, Sarah and Matejovicova, Tatiana and Ram{\'e}, Alexandre and Rivi{\`e}re, Morgane and others},
  journal={arXiv preprint arXiv:2503.19786},
  year={2025}
}

@article{yang2025qwen3,
  title={Qwen3 technical report},
  author={Yang, An and Li, Anfeng and Yang, Baosong and Zhang, Beichen and Hui, Binyuan and Zheng, Bo and Yu, Bowen and Gao, Chang and Huang, Chengen and Lv, Chenxu and others},
  journal={arXiv preprint arXiv:2505.09388},
  year={2025}
}

@misc{nvidia2025nvidianemotronnano2,
      title={NVIDIA Nemotron Nano 2: An Accurate and Efficient Hybrid Mamba-Transformer Reasoning Model}, 
      author={NVIDIA and : and Aarti Basant and Abhijit Khairnar and Abhijit Paithankar and Abhinav Khattar and Adithya Renduchintala and Aditya Malte and Akhiad Bercovich and others},
      year={2025},
      eprint={2508.14444},
      archivePrefix={arXiv},
      primaryClass={cs.CL},
      url={https://arxiv.org/abs/2508.14444}, 
}

@article{he2025skywork,
  title={Skywork Open Reasoner 1 Technical Report},
  author={He, Jujie and Liu, Jiacai and Liu, Chris Yuhao and Yan, Rui and Wang, Chaojie and Cheng, Peng and Zhang, Xiaoyu and Zhang, Fuxiang and Xu, Jiacheng and Shen, Wei and Li, Siyuan and Zeng, Liang and Wei, Tianwen and Cheng, Cheng and An, Bo and Liu, Yang and Zhou, Yahui},
  journal={arXiv preprint arXiv:2505.22312},
  year={2025}
}

@article{yang2025speculative,
  title={Speculative thinking: Enhancing small-model reasoning with large model guidance at inference time},
  author={Yang, Wang and Yue, Xiang and Chaudhary, Vipin and Han, Xiaotian},
  journal={arXiv preprint arXiv:2504.12329},
  year={2025}
}

@article{yang2025specexit,
  title={SpecExit: Accelerating Large Reasoning Model via Speculative Exit},
  author={Yang, Rubing and Bai, Huajun and Liu, Song and Yu, Guanghua and Fan, Runzhi and Dang, Yanbin and Zhang, Jiejing and Liu, Kai and Zhu, Jianchen and Chen, Peng},
  journal={arXiv preprint arXiv:2509.24248},
  year={2025}
}

@article{wang2025wait,
  title={Wait, We Don't Need to" Wait"! Removing Thinking Tokens Improves Reasoning Efficiency},
  author={Wang, Chenlong and Feng, Yuanning and Chen, Dongping and Chu, Zhaoyang and Krishna, Ranjay and Zhou, Tianyi},
  journal={arXiv preprint arXiv:2506.08343},
  year={2025}
}

@article{fang2025thinkless,
  title={Thinkless: Llm learns when to think},
  author={Fang, Gongfan and Ma, Xinyin and Wang, Xinchao},
  journal={arXiv preprint arXiv:2505.13379},
  year={2025}
}

@article{gu2024attention,
  title={When attention sink emerges in language models: An empirical view},
  author={Gu, Xiangming and Pang, Tianyu and Du, Chao and Liu, Qian and Zhang, Fengzhuo and Du, Cunxiao and Wang, Ye and Lin, Min},
  journal={arXiv preprint arXiv:2410.10781},
  year={2024}
}

@article{yue2025don,
  title={Don't Overthink It: A Survey of Efficient R1-style Large Reasoning Models},
  author={Yue, Linan and Du, Yichao and Wang, Yizhi and Gao, Weibo and Yao, Fangzhou and Wang, Li and Liu, Ye and Xu, Ziyu and Liu, Qi and Di, Shimin and others},
  journal={arXiv preprint arXiv:2508.02120},
  year={2025}
}

@inproceedings{zhang2025beyond,
  title={Beyond gpt-5: Making llms cheaper and better via performance-efficiency optimized routing},
  author={Zhang, Yiqun and Li, Hao and Chen, Jianhao and Zhang, Hangfan and Ye, Peng and Bai, Lei and Hu, Shuyue},
  booktitle={Proceedings of the 2025 7th International Conference on Distributed Artificial Intelligence},
  pages={122--129},
  year={2025}
}

@article{liu2025reinforcement,
  title={Reinforcement learning meets large language models: A survey of advancements and applications across the llm lifecycle},
  author={Liu, Keliang and Yang, Dingkang and Qian, Ziyun and Yin, Weijie and Wang, Yuchi and Li, Hongsheng and Liu, Jun and Zhai, Peng and Liu, Yang and Zhang, Lihua},
  journal={arXiv preprint arXiv:2509.16679},
  year={2025}
}

@article{jiang2025think,
  title={Think only when you need with large hybrid-reasoning models},
  author={Jiang, Lingjie and Wu, Xun and Huang, Shaohan and Dong, Qingxiu and Chi, Zewen and Dong, Li and Zhang, Xingxing and Lv, Tengchao and Cui, Lei and Wei, Furu},
  journal={arXiv preprint arXiv:2505.14631},
  year={2025}
}

@article{yang2026ace,
  title={ACE-Bench: Agent Configurable Evaluation with Scalable Horizons and Controllable Difficulty under Lightweight Environments},
  author={Yang, Wang and Song, Chaoda and Li, Xinpeng and Ganguly, Debargha and Ma, Chuang and Wang, Shouren and Dou, Zhihao and Zhou, Yuli and Chaudhary, Vipin and Han, Xiaotian},
  journal={arXiv preprint arXiv:2604.06111},
  year={2026}
}

@article{ganguly2026trust,
  title={Trust The Typical},
  author={Ganguly, Debargha and Sankar, Sreehari and Zhang, Biyao and Singh, Vikash and Gupta, Kanan and Kavuru, Harshini and Luo, Alan and Chen, Weicong and Morningstar, Warren and Machiraju, Raghu and others},
  journal={arXiv preprint arXiv:2602.04581},
  year={2026}
}

@article{wang2025demystifying,
  title={Demystifying Hybrid Thinking: Can LLMs Truly Switch Between Think and No-Think?},
  author={Wang, Shouren and Yang, Wang and Long, Xianxuan and Wang, Qifan and Chaudhary, Vipin and Han, Xiaotian},
  journal={arXiv preprint arXiv:2510.12680},
  year={2025}
}

@inproceedings{yangspeculative,
  title={Speculative Thinking: Enhancing Small-Model Reasoning with Large Model Guidance at Inference Time},
  author={Yang, Van and Yue, Xiang and Chaudhary, Vipin and Han, Xiaotian},
  booktitle={Second Conference on Language Modeling},
  year={2025}
}

@article{yang2025longer,
  title={Longer context, deeper thinking: Uncovering the role of long-context ability in reasoning},
  author={Yang, Wang and Liu, Zirui and Jin, Hongye and Yin, Qingyu and Chaudhary, Vipin and Han, Xiaotian},
  journal={arXiv preprint arXiv:2505.17315},
  year={2025}
}

@inproceedings{yang2025100,
  title={100-LongBench: Are de facto Long-Context Benchmarks Literally Evaluating Long-Context Ability?},
  author={Yang, Van and Jin, Hongye and Zhong, Shaochen and Jiang, Song and Wang, Qifan and Chaudhary, Vipin and Han, Xiaotian},
  booktitle={Findings of the Association for Computational Linguistics: ACL 2025},
  pages={17560--17576},
  year={2025}
}

@article{jain2024livecodebench,
  title={Livecodebench: Holistic and contamination free evaluation of large language models for code},
  author={Jain, Naman and Han, King and Gu, Alex and Li, Wen-Ding and Yan, Fanjia and Zhang, Tianjun and Wang, Sida and Solar-Lezama, Armando and Sen, Koushik and Stoica, Ion},
  journal={arXiv preprint arXiv:2403.07974},
  year={2024}
}

@article{xu2025towards,
  title={Towards large reasoning models: A survey of reinforced reasoning with large language models},
  author={Xu, Fengli and Hao, Qianyue and Zong, Zefang and Wang, Jingwei and Zhang, Yunke and Wang, Jingyi and Lan, Xiaochong and Gong, Jiahui and Ouyang, Tianjian and Meng, Fanjin and others},
  journal={arXiv preprint arXiv:2501.09686},
  year={2025}
}

@article{xiao2023efficient,
  title={Efficient streaming language models with attention sinks},
  author={Xiao, Guangxuan and Tian, Yuandong and Chen, Beidi and Han, Song and Lewis, Mike},
  journal={arXiv preprint arXiv:2309.17453},
  year={2023}
}

@article{zhang2025survey,
  title={A survey of reinforcement learning for large reasoning models},
  author={Zhang, Kaiyan and Zuo, Yuxin and He, Bingxiang and Sun, Youbang and Liu, Runze and Jiang, Che and Fan, Yuchen and Tian, Kai and Jia, Guoli and Li, Pengfei and others},
  journal={arXiv preprint arXiv:2509.08827},
  year={2025}
}

@inproceedings{robinson2024sparse,
  title={A sparse null code emerges in deep neural networks},
  author={Robinson, Brian S and Drenkow, Nathan and Conwell, Colin and Bonner, Michael},
  booktitle={Proceedings of UniReps: The First Workshop on Unifying Representations in Neural Models},
  pages={302--314},
  year={2024},
  organization={PMlR}
}

@article{sun2024massive,
  title={Massive activations in large language models},
  author={Sun, Mingjie and Chen, Xinlei and Kolter, J Zico and Liu, Zhuang},
  journal={arXiv preprint arXiv:2402.17762},
  year={2024}
}

@article{zhang2025interplay,
  title={On the Interplay of Pre-Training, Mid-Training, and RL on Reasoning Language Models},
  author={Zhang, Charlie and Neubig, Graham and Yue, Xiang},
  journal={arXiv preprint arXiv:2512.07783},
  year={2025}
}

@article{xi2025bapo,
  title={BAPO: Stabilizing Off-Policy Reinforcement Learning for LLMs via Balanced Policy Optimization with Adaptive Clipping},
  author={Xi, Zhiheng and Guo, Xin and Nan, Yang and Zhou, Enyu and Shen, Junrui and Chen, Wenxiang and Liu, Jiaqi and Huang, Jixuan and Zhang, Zhihao and Guo, Honglin and others},
  journal={arXiv preprint arXiv:2510.18927},
  year={2025}
}

@article{kang2025quagmires,
  title={Quagmires in SFT-RL Post-Training: When High SFT Scores Mislead and What to Use Instead},
  author={Kang, Feiyang and Kuchnik, Michael and Padthe, Karthik and Vlastelica, Marin and Jia, Ruoxi and Wu, Carole-Jean and Ardalani, Newsha},
  journal={arXiv preprint arXiv:2510.01624},
  year={2025}
}

@article{wu2025takes,
  title={It Takes Two: Your GRPO Is Secretly DPO},
  author={Wu, Yihong and Ma, Liheng and Ding, Lei and Li, Muzhi and Wang, Xinyu and Chen, Kejia and Su, Zhan and Zhang, Zhanguang and Huang, Chenyang and Zhang, Yingxue and others},
  journal={arXiv preprint arXiv:2510.00977},
  year={2025}
}

@article{hao2025rethinking,
  title={Rethinking entropy interventions in rlvr: An entropy change perspective},
  author={Hao, Zhezheng and Wang, Hong and Liu, Haoyang and Luo, Jian and Yu, Jiarui and Dong, Hande and Lin, Qiang and Wang, Can and Chen, Jiawei},
  journal={arXiv preprint arXiv:2510.10150},
  year={2025}
}

@article{nan2025ngrpo,
  title={Ngrpo: Negative-enhanced group relative policy optimization},
  author={Nan, Gongrui and Chen, Siye and Huang, Jing and Lu, Mengyu and Wang, Dexun and Xie, Chunmei and Xiong, Weiqi and Zeng, Xianzhou and Zhou, Qixuan and Li, Yadong and others},
  journal={arXiv preprint arXiv:2509.18851},
  year={2025}
}

@article{zheng2025group,
  title={Group sequence policy optimization},
  author={Zheng, Chujie and Liu, Shixuan and Li, Mingze and Chen, Xiong-Hui and Yu, Bowen and Gao, Chang and Dang, Kai and Liu, Yuqiong and Men, Rui and Yang, An and others},
  journal={arXiv preprint arXiv:2507.18071},
  year={2025}
}

@misc{skywork-or1-2025,
  title={Skywork Open Reasoner Series},
  author = {He, Jujie and Liu, Jiacai and Liu, Chris Yuhao and Yan, Rui and Wang, Chaojie and Cheng, Peng and Zhang, Xiaoyu and Zhang, Fuxiang and Xu, Jiacheng and Shen, Wei and Li, Siyuan and Zeng, Liang and Wei, Tianwen and Cheng, Cheng and Liu, Yang and Zhou, Yahui},
  
  note={Notion Blog},
  year={2025}
}

@article{xu2025scalable,
  title={Scalable chain of thoughts via elastic reasoning},
  author={Xu, Yuhui and Dong, Hanze and Wang, Lei and Sahoo, Doyen and Li, Junnan and Xiong, Caiming},
  journal={arXiv preprint arXiv:2505.05315},
  year={2025}
}

@article{xu2025thinking,
  title={Thinking-Free Policy Initialization Makes Distilled Reasoning Models More Effective and Efficient Reasoners},
  author={Xu, Xin and AI, Cliveb and Yang, Kai and Chen, Tianhao and Wang, Yang and Yang, Saiyong and Yang, Can},
  journal={arXiv preprint arXiv:2509.26226},
  year={2025}
}

@article{moshkov2025aimo2,
  title   = {AIMO-2 Winning Solution: Building State-of-the-Art Mathematical Reasoning Models with OpenMathReasoning dataset},
  author  = {Ivan Moshkov and Darragh Hanley and Ivan Sorokin and Shubham Toshniwal and Christof Henkel and Benedikt Schifferer and Wei Du and Igor Gitman},
  year    = {2025},
  journal = {arXiv preprint arXiv:2504.16891}
}

\newpage
\appendix

\section{Appendix}
\label{sec:appendix}

\subsection{Quantitative Attention Analysis}
\label{sec:quant_attn}
To further quantify this observation, we compute the \textit{trigger-token attention mass} for each reasoning mode on Qwen3-8B.
Specifically, we measure the average attention assigned to trigger tokens relative to all other tokens in the prompt, and define the ratio as:
\begin{equation}
    \text{Ratio} = \frac{\text{Attention on Trigger Token}}{\text{Avg Attention on Other Tokens}}
\end{equation}
\Cref{tab:attn_mass} reports these ratios alongside MATH-500 accuracy and average generation length.
In No-Think mode, the post-\texttt{</think>} newline token dominates with a ratio of 5.27$\times$, suppressing reasoning.
In Think mode, the ``\texttt{Okay}'' token dominates with a 3.18$\times$ ratio, activating reasoning.
Crucially, Mid-Think simultaneously activates \emph{both} trigger regions: ``\texttt{Okay}'' at 2.10$\times$ and the newline at 3.08$\times$, resulting in intermediate accuracy (92.3\%) and generation length between the two extremes.
This quantitative evidence confirms that Mid-Think induces a genuinely hybrid attention pattern, rather than simply averaging the two modes.

\begin{table}[ht]
\centering
\small
\begin{tabular}{llccc}
\toprule
\textbf{Mode} & \textbf{Trigger} & \textbf{Ratio} & \textbf{Acc (\%)} & \textbf{Avg Len} \\
\midrule
No-Think  & \texttt{\textbackslash n\textbackslash n}  & 5.27$\times$ & 83.2 & 1{,}012 \\
Mid-Think & \texttt{\textbackslash n\textbackslash n}  & 3.08$\times$ & 92.3 & 2{,}753 \\
Mid-Think & \texttt{Okay}                              & 2.10$\times$ & 92.3 & 2{,}753 \\
Think     & \texttt{Okay}                              & 3.18$\times$ & 94.6 & 5{,}557 \\
\bottomrule
\end{tabular}
\caption{Trigger-token attention mass ratio, MATH-500 accuracy, and average generation length under different reasoning modes on Qwen3-8B. Mid-Think jointly activates both trigger regions, yielding hybrid behavior.}
\label{tab:attn_mass}
\end{table}

\subsection{Another Verification of Budget-Controlled Reasoning Baseline}


Previously, we showed that the proposed budget-controlled reasoning method enables proportional control over the reasoning budget, resulting in systematic changes in average output length, accuracy, and wait count. In this section, we present the corresponding quantitative results. As shown in Tables~\ref{tab:budget_verification_math500} and~\ref{tab:budget_verification_math500_qwen3_8b}, all three metrics increase monotonically with the reasoning budget, consistently exhibiting the expected budget–performance trade-off.

\begin{table}[ht]
\centering
\small
\setlength{\tabcolsep}{8pt}
\begin{tabular}{c c c c}
\toprule
\textbf{Budget} & \textbf{Avg Length} & \textbf{Wait} & \textbf{Acc (\%)} \\
\midrule
0.0 (No-think) &  899.1 &   329 & 86.3 \\
0.1 & 1563.3 & 10252 & 87.4 \\
0.2 & 1879.9 & 16225 & 89.2 \\
0.3 & 2170.8 & 22293 & 91.6 \\
0.4 & 2472.7 & 27339 & 92.2 \\
0.5 & 2852.9 & 33080 & 92.8 \\
0.6 & 3244.6 & 38788 & 93.6 \\
1.0 (Think) & 4904.3 & 59288 & 94.4 \\
\bottomrule
\end{tabular}
\vspace{-8pt}
\caption{
Verification of budget-controlled method on MATH500 with Qwen3-14B. We report average output length, wait count, and accuracy under different budgets. As the budget increases, all three metrics improve steadily, demonstrating the effectiveness of the method.
}
\label{tab:budget_verification_math500}
\end{table}

\begin{table}[ht]
\centering
\small
\setlength{\tabcolsep}{8pt}
\begin{tabular}{c c c c}
\toprule
\textbf{Budget} & \textbf{Avg Length} & \textbf{Wait} & \textbf{Acc (\%)} \\
\midrule
0.0 (No-think) & 1012.6 &  2082  & 83.2 \\
0.1 & 2145.3 & 19679 & 88.0 \\
0.2 & 2466.6 & 26778 & 89.8 \\
0.3 & 2709.0 & 33543 & 90.3 \\
0.4 & 3085.2 & 40188 & 92.0 \\
0.5 & 3456.7 & 52871 & 92.8 \\
0.6 & 3788.3 & 53260 & 93.3 \\
1.0 (Think) & 5557.4 & 81367 & 94.6 \\
\bottomrule
\end{tabular}
\vspace{-8pt}
\caption{
Verification of budget-controlled reasoning on \texttt{MATH500} using \texttt{Qwen3-8B}. We report average output length, wait count, and accuracy under different budgets. As the reasoning budget increases, all three metrics improve steadily, demonstrating effective budget control.
}
\label{tab:budget_verification_math500_qwen3_8b}
\end{table}

\subsection{Results of Mid-Think v.s. Different Budget}


Previously, we presented the performance of Mid-Think across different model scales primarily through visualizations. To provide more concrete evidence, we now report the corresponding quantitative results. Tables~\ref{tab:budget_verification_gpqa_qwen3_14b}, \ref{tab:budget_verification_aime_qwen3_14b}, and \ref{tab:budget_verification_math500_qwen3_14b_full} summarize the detailed numerical results. Across all datasets, Mid-Think—regardless of the specific tag used—consistently achieves intermediate-budget reasoning behavior. Notably, on GPQA, certain Mid-Think variants even outperform standard budget scaling under comparable or lower reasoning cost.

\begin{table}[ht]
\centering
\small
\setlength{\tabcolsep}{7pt}
\begin{tabular}{c c c c c}
\toprule
\textbf{Format} & \textbf{Budget} & \textbf{Avg Length} & \textbf{Wait} & \textbf{Acc} \\
\midrule
standard & 0.0 &  899.1 &   329  & 86.3 \\
standard & 0.1 & 1563.3 & 10252 & 87.4 \\
standard & 0.2 & 1879.9 & 16225 & 89.2 \\
standard & 0.3 & 2170.8 & 22293 & 91.6 \\
standard & 0.4 & 2472.7 & 27339 & 92.2 \\
standard & 0.5 & 2852.9 & 33080 & 92.8 \\
standard & 0.6 & 3244.6 & 38788 & 93.6 \\
standard & 1.0 & 4904.3 & 59288 & 94.4 \\
\midrule
begin & 1.0 & 2805.5 & 23024 & 93.1 \\
reason & 1.0 & 2589.8 & 20933 & 92.1 \\
less\_think & 1.0 & 2655.0 & 21249 & 93.3 \\
\bottomrule
\end{tabular}
\vspace{-8pt}
\caption{
Quantitative results of budget-controlled reasoning on \texttt{MATH500} using \texttt{Qwen3-14B}. We report average output length, wait count, and accuracy under different reasoning budgets and prompt formats. Standard budget control exhibits a clear monotonic trade-off between budget and performance, while alternative formats (\texttt{begin}, \texttt{reason}, \texttt{less\_think}) achieve comparable accuracy with substantially reduced reasoning length.
}
\label{tab:budget_verification_math500_qwen3_14b_full}
\end{table}

\begin{table}[ht]
\centering
\small
\setlength{\tabcolsep}{7pt}
\begin{tabular}{c c c c c}
\toprule
\textbf{Format} & \textbf{Budget} & \textbf{Avg Length} & \textbf{Wait} & \textbf{Acc} \\
\midrule
standard & 0.0 & 4084.9 &   501  & 22.9 \\
standard & 0.1 & 5884.0 &  7070 & 33.1 \\
standard & 0.2 & 6515.3 & 10911 & 37.8 \\
standard & 0.3 & 7784.5 & 14455 & 42.9 \\
standard & 0.4 & 8626.4 & 17597 & 48.9 \\
standard & 0.5 & 9449.8 & 20143 & 54.7 \\
standard & 0.6 & 10486.8 & 21793 & 57.6 \\
standard & 1.0 & 15792.0 & 31686 & 74.4 \\
\midrule
begin & 1.0 & 11717.4 & 17083 & 61.3 \\
reason & 1.0 & 10862.2 & 14660 & 58.7 \\
less\_think & 1.0 & 10747.9 & 14893 & 61.1 \\
\bottomrule
\end{tabular}
\vspace{-8pt}
\caption{
Results of budget-controlled reasoning on \texttt{AIME} using \texttt{Qwen3-14B}. We report average output length, wait count, and accuracy under different reasoning budgets and prompt formats. Increasing the budget leads to substantial performance gains, while alternative formats (\texttt{begin}, \texttt{reason}, \texttt{less\_think}) achieve competitive accuracy with reduced reasoning cost compared to standard full-thinking.
}
\label{tab:budget_verification_aime_qwen3_14b}
\end{table}

\begin{table}[ht]
\centering
\small
\setlength{\tabcolsep}{7pt}
\begin{tabular}{c c c c c}
\toprule
\textbf{Format} & \textbf{Budget} & \textbf{Avg Length} & \textbf{Wait} & \textbf{Acc} \\
\midrule
standard & 0.0 & 1211.4 &   398  & 39.4 \\
standard & 0.1 & 1756.2 &  7836 & 41.5 \\
standard & 0.2 & 2351.2 & 13849 & 41.7 \\
standard & 0.3 & 3005.8 & 20242 & 42.0 \\
standard & 0.4 & 3606.8 & 25896 & 47.7 \\
standard & 0.5 & 4208.1 & 30162 & 46.9 \\
standard & 0.6 & 5000.8 & 35780 & 47.4 \\
standard & 1.0 & 7799.4 & 50661 & 63.1 \\
\midrule
begin & 1.0 & 2160.2 &  6138 & 50.4 \\
reason & 1.0 & 1763.0 &  4159 & 53.9 \\
less\_think & 1.0 & 2073.1 &  4999 & 50.4 \\
\bottomrule
\end{tabular}
\vspace{-8pt}
\caption{
Results of budget-controlled reasoning on \texttt{GPQA} using \texttt{Qwen3-14B}. We report average output length, wait count, and accuracy under different reasoning budgets and prompt formats. Standard budget control shows increasing performance with higher budgets, while alternative formats achieve competitive accuracy with substantially reduced reasoning cost.
}
\label{tab:budget_verification_gpqa_qwen3_14b}
\end{table}

\subsection{Comparison Results on Qwen3-4B}


Previously, we focused our analysis on hybrid models at the 8B and 14B scales. Here, we additionally report results on \texttt{Qwen3-4B}. As shown in Figure~\ref{fig:token_cue_all}, Mid-Think continues to induce intermediate-budget reasoning behavior on \texttt{Qwen3-4B}, demonstrating that the effect generalizes to smaller model scales.

\begin{figure*}[t]
    \centering
    \begin{subfigure}[t]{0.32\textwidth}
        \centering
        \includegraphics[width=\linewidth]{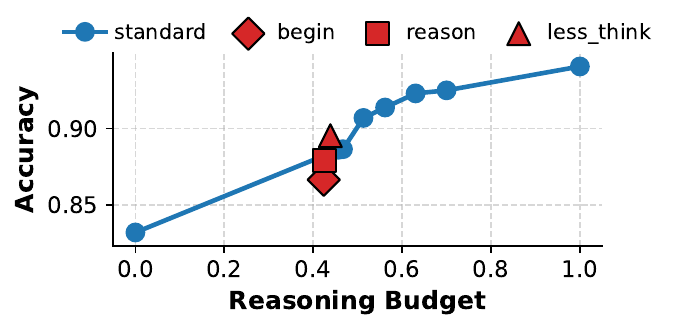}
        \caption{Qwen3-4B on MATH500}
    \end{subfigure}
    \hfill
    \begin{subfigure}[t]{0.32\textwidth}
        \centering
        \includegraphics[width=\linewidth]{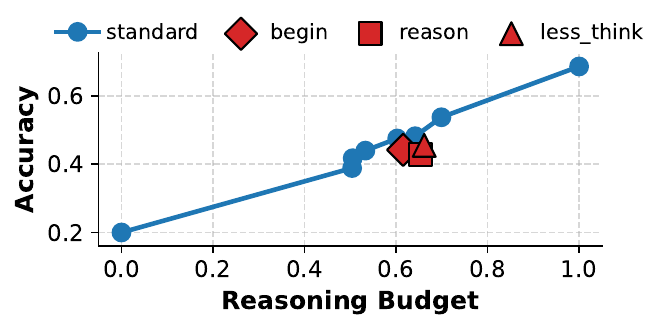}
        \caption{Qwen3-4B on AIME}
    \end{subfigure}
    \hfill
    \begin{subfigure}[t]{0.32\textwidth}
        \centering
        \includegraphics[width=\linewidth]{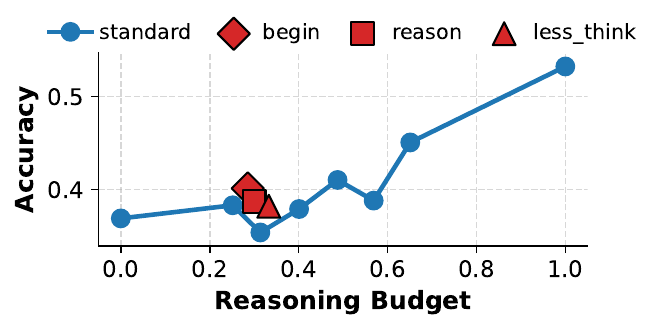}
        \caption{Qwen3-4B on GPQA}
    \end{subfigure}

    \vspace{0.8em}
    \caption{
        Comparison between different budget reasoning and Mid-Think on pure-think and RL-based models across multiple datasets. We report results for Qwen3-4B under varying reasoning budgets, together with Mid-Think using different tags (<reason>, \texttt{<begin>}, \texttt{<less think>}). Across MATH500, AIME, and GPQA, Mid-Think consistently achieves performance corresponding to intermediate reasoning budgets, and on GPQA it surpasses fixed-budget baselines, yielding Pareto-optimal accuracy--efficiency trade-offs between Think and No-Think mode.
        }
    \label{fig:token_cue_3x3 Qwen3-4B}
\end{figure*}

\begin{table}[t]
\centering
\small
\setlength{\tabcolsep}{6pt}
\begin{tabular}{c c c c}
\toprule
\textbf{Training} & \textbf{Mode} & \textbf{Test} & \textbf{Acc / Len} \\
\midrule

NO & --
& No-think
& 83.2 / 1013 \\
& 
& Think
& 94.6 / 5557 \\

\midrule

RL & Think
& No-think
& 81.9 / 1392 \\
& 
& Think
& 93.6 / 4568 \\

\midrule

RL & No-Think
& No-think
& 91.6 / 2268 \\
& 
& Think
& 93.6 / 6335 \\

\midrule

RL & Mid-Think
& No-think
& 85.8 / 1374 \\
& 
& Think
& \textbf{94.1} / 5302 \\

\bottomrule
\end{tabular}
\caption{
Performance of \texttt{Qwen3-8B} under different training regimes.
We report accuracy (\%) and average generation length.
Each trained model is evaluated under both No-think and Think test settings.
Reason (Mid-Think) training achieves the best Think-Test accuracy with balanced generation length.
}
\label{tab:qwen3_8b_train_modes}
\end{table}

\subsection{Mid-Think to RL Training after SFT on MATH500}


Previously, we only reported results on AIME.
We now extend our evaluation to \textbf{Math500} and examine the effect of applying \textbf{Mid-Think} as the reinforcement learning objective.
Across both \texttt{Qwen3-8B} and \texttt{Qwen3-4B}, Mid-Think training consistently improves performance under the \emph{Think} test setting, while largely preserving behavior under the \emph{No-think} test setting.

As shown in Tables~\ref{tab:qwen3_8b_train_modes} and~\ref{tab:qwen3_4b_math500_train_modes}, models trained with standard Think supervision remain inferior to those trained with Mid-Think when evaluated in Think mode.
Meanwhile, Mid-Think training maintains competitive accuracy under No-think evaluation, indicating that it serves as an effective intermediate reasoning objective that balances reasoning strength and inference efficiency.

\begin{table}[ht]
\centering
\small
\setlength{\tabcolsep}{6pt}
\begin{tabular}{c c c c}
\toprule
\textbf{Training} & \textbf{Mode} & \textbf{Test} & \textbf{Acc / Len} \\
\midrule

No & --
& No-think
& 83.2 / 991 \\
& 
& Think
& 94.0 / 5334 \\

\midrule

RL & Think
& No-think
& 80.8 / 1362 \\
& 
& Think
& 92.4 / 4061 \\

\midrule

RL & No-Think
& No-think
& 88.4 / 3234 \\
& 
& Think
& 93.9 / 7458 \\

\midrule

RL & Mid-Think
& No-think
& 83.9 / 1852 \\
& 
& Think
& 93.9 / 5456 \\

\bottomrule
\end{tabular}
\caption{
Performance of \texttt{Qwen3-4B} on \textbf{Math500} under different training regimes.
We report accuracy (\%) and average generation length.
Each model is evaluated under both No-think and Think test settings.
Mid-Think achieves strong Think-Test accuracy while better preserving No-think behavior compared to standard Think training.
}
\label{tab:qwen3_4b_math500_train_modes}
\end{table}

\subsection{Training Hyperparameters}
\label{sec:hyperparams}

To facilitate reproducibility, we summarize the full set of training hyperparameters in \Cref{tab:hyperparams}. Our setup follows the official VERL GRPO recipe, and all three training modes (Think, No-Think, and Mid-Think) use identical hyperparameters to ensure a fair comparison.

\paragraph{Response sampling.}
We sample 8 responses per prompt during rollout, which provides sufficient diversity for advantage estimation in GRPO while remaining computationally tractable. The maximum response length is capped at 16,384 tokens to accommodate long chain-of-thought outputs under the Think setting, while also allowing Mid-Think responses to naturally vary in length without artificial truncation.

\paragraph{Training data and epochs.}
We train on a curated set of 5,000 problems for 6 epochs. This relatively compact dataset size is intentional: it reduces the risk of reward hacking on easy samples while maintaining a sufficient diversity of problem types. Training for 6 epochs strikes a balance between convergence and overfitting, as we observed diminishing returns beyond this point in preliminary experiments.

\paragraph{Optimization settings.}
We use a learning rate of \(1 \times 10^{-6}\) with a warmup ratio of 0.05 and weight decay of 0.1. The small learning rate is chosen conservatively to preserve the pretrained model's language generation capabilities while allowing the policy to shift meaningfully under RL. The weight decay serves as a regularizer to prevent large deviations from the reference policy, complementing the KL penalty already present in GRPO.

\paragraph{Batch configuration.}
The train batch size is 256, decomposed into PPO mini-batches of 64, with a micro-batch size of 16 per GPU. This hierarchical batching strategy is standard in large-scale RLHF pipelines and allows gradient accumulation across multiple forward passes while keeping per-device memory usage manageable.

\begin{table}[t]
\centering
\small
\begin{tabular}{lc}
\toprule
\textbf{Hyperparameter} & \textbf{Value} \\
\midrule
Responses per prompt         & 8 \\
Max response length (tokens) & 16{,}384 \\
Training epochs              & 6 \\
Training data size           & 5{,}000 \\
Learning rate                & 1e-6 \\
LR warmup ratio              & 0.05 \\
Weight decay                 & 0.1 \\
Train batch size             & 256 \\
PPO mini-batch size          & 64 \\
Micro-batch size per GPU     & 16 \\
\bottomrule
\end{tabular}
\caption{Training hyperparameters for GRPO-based RL experiments. All three training modes use identical settings.}
\label{tab:hyperparams}
\end{table}

\end{document}